\documentclass{article}
\usepackage[T1]{fontenc}
\usepackage{longcat}
\usepackage{iftex}
\ifPDFTeX
\fi
\usepackage[authoryear,round]{natbib}
\setcitestyle{authoryear,round,citesep={;},aysep={,},yysep={;}}
\renewcommand{\shorttitle}{N-OPSD}
\renewcommand{\headeright}{}

\usepackage{amsmath,amsfonts,bm}

\def\eqref#1{equation~\ref{#1}}

\def\1{\bm{1}}

\DeclareMathAlphabet{\mathsfit}{\encodingdefault}{\sfdefault}{m}{sl}
\SetMathAlphabet{\mathsfit}{bold}{\encodingdefault}{\sfdefault}{bx}{n}

\usepackage{amsmath}
\usepackage{amssymb}
\usepackage{booktabs}
\usepackage{xcolor,colortbl,tabularx}
\usepackage{graphicx}
\usepackage{placeins}
\usepackage{multirow}
\usepackage{hyperref}
\usepackage{url}
\hypersetup{
  pdftitle={Better Supervision Is Nearby: Neighborhood On-Policy Self-Distillation}
}

\newcommand{\method}{\textsc{Neighborhood OPSD}}
\newcommand{\methodshort}{\textsc{N-OPSD}}
\newcommand{\mzero}{\ensuremath{M_0}}

\title{Better Supervision Is Nearby:\\
Neighborhood On-Policy Self-Distillation}

\author{%
  {\large
    \textbf{Xincheng Wei}\textsuperscript{1,*}\qquad
    \textbf{Yifan Ding}\textsuperscript{2,*}\qquad
    \textbf{Yoshua Li}\textsuperscript{2,*,\textdagger}
  }\\[0.5em]
  {\large
    \textbf{Yuquan Lu}\textsuperscript{2}\qquad
    \textbf{Ziheng Li}\textsuperscript{2}\qquad
    \textbf{Yi Lu}\textsuperscript{2,4}
  }\\[0.5em]
  {\large
    \textbf{Dongsheng Ma}\textsuperscript{2,3}\qquad
    \textbf{Rongxiang Weng}\textsuperscript{2}\qquad
    \textbf{Xunliang Cai}\textsuperscript{2}
  }\\[0.85em]
  {\normalsize
    \textsuperscript{1}The Chinese University of Hong Kong, Shenzhen\\[0.2em]
    \textsuperscript{2}Meituan, LongCat Team\\[0.2em]
    \textsuperscript{3}Peking University\qquad
    \textsuperscript{4}University of Toronto
  }\\[0.65em]
  {\small
    \textsuperscript{*}Equal contribution.\qquad
    \textsuperscript{\textdagger}Corresponding author.\\[0.2em]
    \texttt{yoshua\_li@meituan.com}
  }
}
\hypersetup{pdfauthor={Xincheng Wei, Yifan Ding, Yoshua Li, Yuquan Lu, Ziheng Li, Yi Lu, Dongsheng Ma, Rongxiang Weng, Xunliang Cai}}

\begin{document}

\maketitle

\begin{abstract}
On-policy self-distillation (OPSD) trains mathematical reasoning models using a
privileged teacher that sees a reference solution and supervises student-sampled
prefixes. Standard OPSD uses one fixed parameter setting at every state, but
nearby settings may offer additional supervision.
We find that local parameter perturbations reveal complementary reference-aligned
corrections under the same reference context. Different experts supply these
corrections at different reference positions. Their pool covers more such
positions than the unperturbed privileged teacher.
We introduce \method{} (\methodshort{}) to turn these corrections
into supervision at student-visited states.
Offline, greedy selection builds a compact pool of frozen experts by rewarding
filtered reference-token gains beyond the pool's current best at each position.
The highest-peak expert need not provide the best training target.
Online routing therefore separates the anchor direction from its level of
support. MaxPeak selects the anchor token, and quantile selection chooses among
experts whose top token matches it. The student learns from the chosen expert's full next-token
distribution through the clipped forward-KL objective inherited from OPSD.
We evaluate on AIME 2024, AIME 2025, and HMMT February 2025.
Across three independent runs per method, \method{} improves the three-benchmark
Average@12 over OPSD by $2.75$, $1.67$, and $1.94$ points on Qwen3-1.7B, 4B,
and 8B, respectively.
Student-prefix continuations support using the pool beyond the reference
trajectories used for selection. Matched ablations support filtered
reference-token gains as a selection criterion. Accounting for overlap within
the pool and routing by state further improve student accuracy.
Inference uses only the distilled student.
\end{abstract}

\section{Introduction}
\label{sec:introduction}

On-policy self-distillation (OPSD) trains mathematical reasoning models using
two copies of the same model \citep{zhao2026selfdistilled}. The student generates
a response from the problem alone. The privileged teacher also sees a reference
solution, available only during training \citep{lopezpaz2016privileged}, and
provides next-token distributions along the student's response. Standard OPSD
uses the same unperturbed teacher
\mzero{} at every student-visited state. Its predictions depend on the student
prefix, but all supervision comes from one fixed parameter setting.
A fixed teacher need not exhaust the useful supervision available from the
same base model. We study its local parameter neighborhood as a source of
complementary corrections for on-policy distillation.

TrustMOPD weights specialist supervision at each student prefix
\citep{sun2026trustmopd}. RandOpt selects local perturbation experts by task
performance \citep{gan2026neuralthickets}. For distillation, an expert's task
accuracy need not reflect the value of its supervision
\citep{cho2019efficacy}.
Task accuracy also does not reveal which corrections the selected experts
already supply. This motivates selecting local experts by the token-level
corrections they add to a pool.

\begin{figure}[t]
    \centering
    \includegraphics[width=\linewidth]{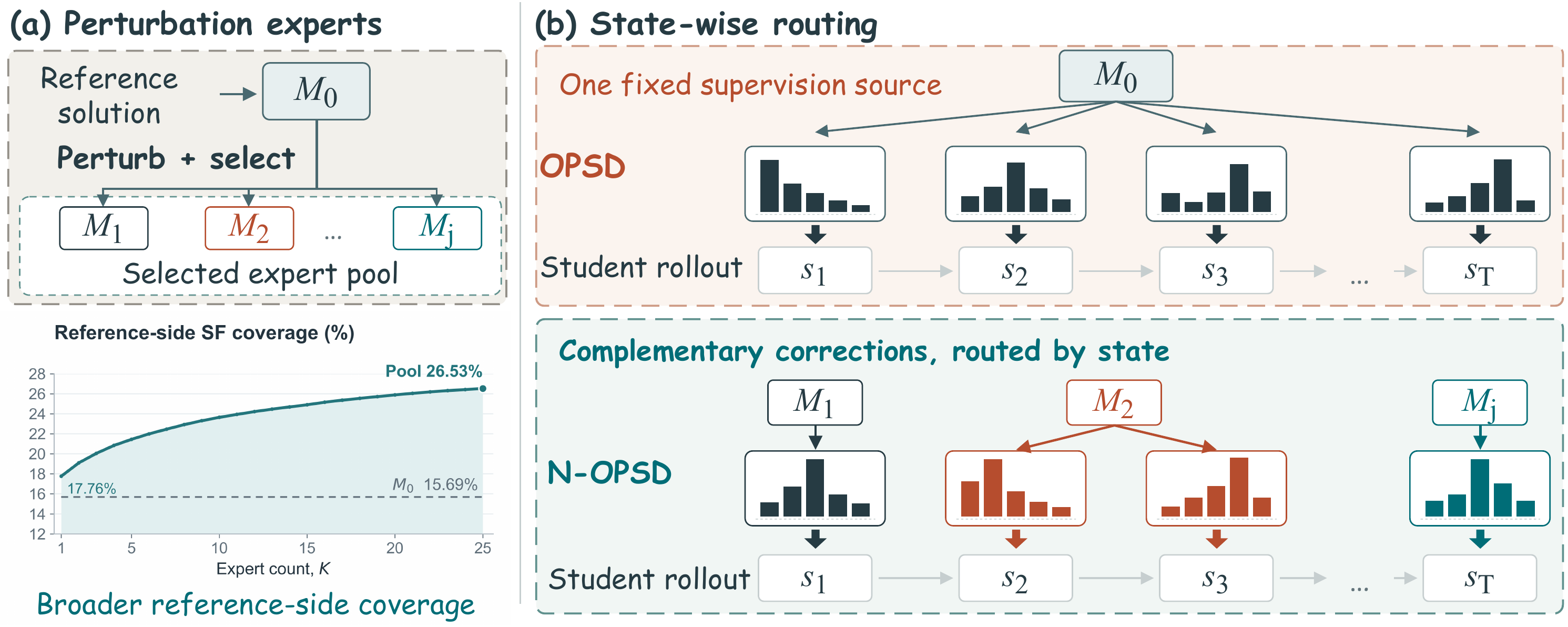}
    \caption{Overview of \method{}.
    (a) Selected local perturbation experts expand reference-side
    selection-feasible (SF) coverage beyond \mzero{} (Qwen3-8B,
    $\sigma=.002$; 500-question analysis set).
    (b) OPSD uses \mzero{} at every student-visited state, while \method{}
    routes one expert's full next-token distribution at each state.
    Colors distinguish selected experts; distributions are schematic.}
    \label{fig:intro_opsd}
\end{figure}

Experts sharing the same base checkpoint and reference context provide
complementary corrections in our reference-side analysis. We measure each
correction as an increase in reference-token probability over the frozen
problem-only model on the same reference prefix. Different experts provide
these increases at different positions. Their pool covers more reference
positions than the unperturbed privileged teacher \mzero{} under the same
selection filters. Thus, useful supervision can depend on which nearby
parameter setting interprets the same privileged information.

We introduce \method{} (\methodshort{}) to turn complementary corrections in
the teacher's neighborhood into supervision at student-visited states
(Figure~\ref{fig:intro_opsd}). Offline selection builds a compact pool by
rewarding experts for reference-aligned corrections beyond what the current
pool provides.
During training, student prefixes can differ from the reference solution
\citep{agarwal2024onpolicy}, so the reference token cannot directly guide routing.
The selected experts therefore evaluate the student's current prefix.

Online routing separates the anchor direction from its level of support.
MaxPeak takes the token with the highest probability assigned by any expert as
the anchor, allowing an individual expert to supply the direction without pool
agreement. A large positive gap between teacher and student probabilities can cause
the anchor's loss term to be clipped \citep{zhao2026selfdistilled}. Quantile selection
chooses the level of support among experts whose top token matches the anchor. The student learns
from the chosen expert's full next-token distribution. All experts remain
fixed, and inference uses only the distilled student.

We evaluate on AIME 2024, AIME 2025, and HMMT February 2025.
Across three independent runs per method, \method{} improves the three-benchmark
Average@12 over OPSD by $2.75$, $1.67$, and $1.94$ points on Qwen3-1.7B, 4B,
and 8B. Student-prefix continuations also show that MaxPeak can improve
continuation accuracy over \mzero{}. This supports using the selected pool
beyond the reference trajectories used for expert selection.
Selection ablations show gains from scoring reference-aligned corrections,
with further gains from accounting for overlap within the pool. Routing
ablations then show gains from choosing the supervision source based on
expert predictions at each state.

The contributions are:
\begin{itemize}
    \item We show that a privileged teacher's reference-aligned supervision
    extends beyond its unperturbed parameter setting. Selected local experts
    provide complementary corrections at reference positions beyond those
    covered by \mzero{}.
    \item We introduce \method{}, a framework that turns a privileged
    teacher's neighborhood into state-wise supervision for on-policy
    distillation. It selects complementary local experts and distills
    their supervision into a single student used alone at inference.
    \item Across three independent runs per method, \method{} improves the
    three-benchmark Average@12 over OPSD by $1.67$ to $2.75$ points on
    Qwen3-1.7B, 4B, and 8B. Matched ablations connect these gains to
    complementary expert selection and state-wise routing.
\end{itemize}

\section{Method}
\label{sec:method}

\begin{figure}[t]
    \centering
    \includegraphics[width=\linewidth]{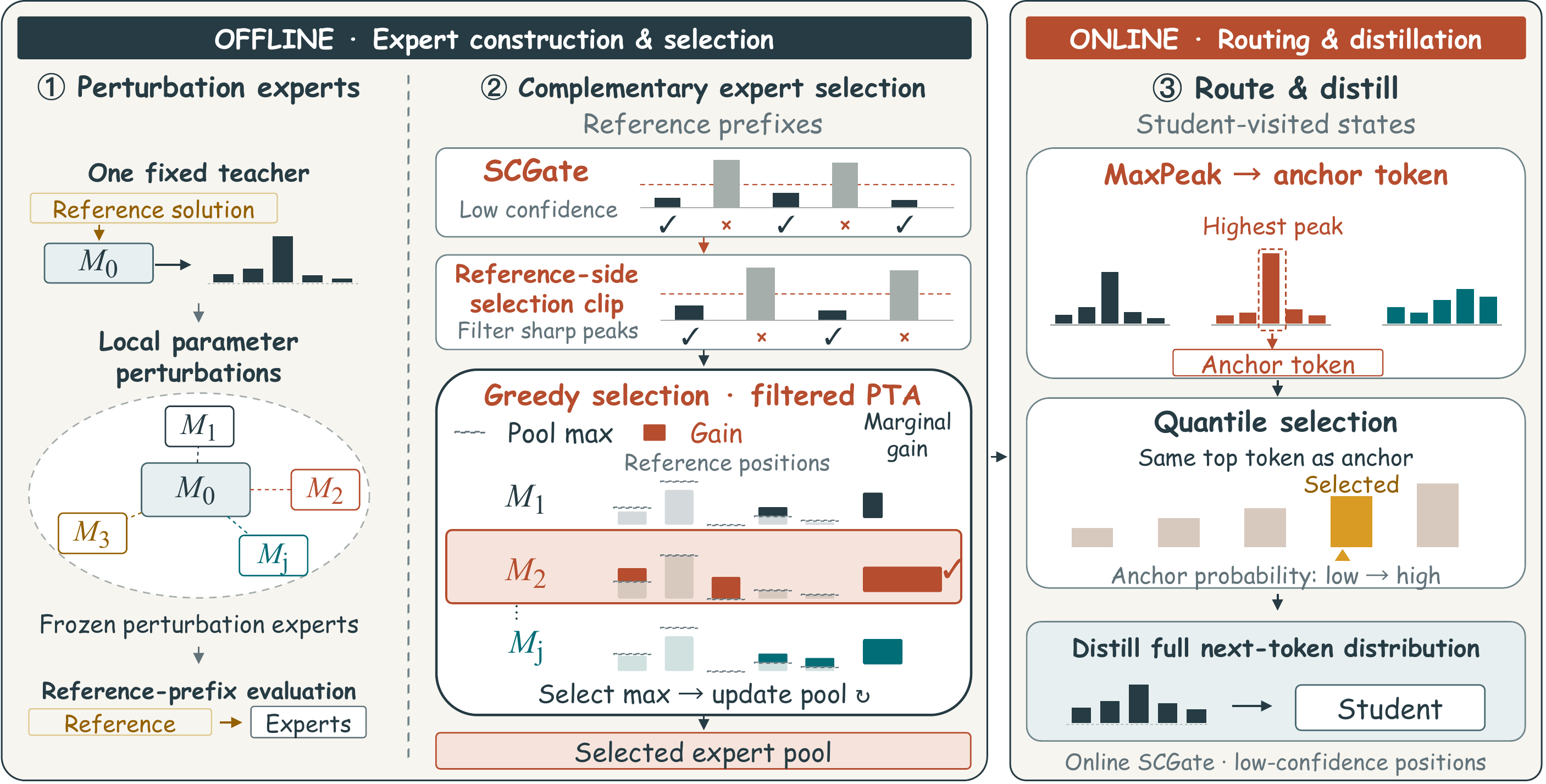}
    \caption{Overview of \method{}.
    Offline, greedy selection builds a complementary pool of local perturbation
    experts by marginal filtered PTA on reference prefixes.
    Online, MaxPeak selects an anchor token, and quantile selection chooses
    among experts whose top token matches it. The selected expert supplies
    the full next-token distribution for student distillation.}
    \label{fig:method_opsd}
\end{figure}

\method{} separates the choice of an expert pool from the choice of a
supervision source at each state (Figure~\ref{fig:method_opsd}).
Offline selection identifies experts that add complementary corrections on
reference prefixes. Online routing then selects from this pool at the
student's current prefix, where the useful supervision source can change
as the rollout develops.

\subsection{Setup and Perturbation Experts}
\label{sec:setup}

Let $x$ be a problem, $z$ a privileged reference solution available only during
training, and $\pi_\theta$ the student. The student samples a response
$\mathbf y=(y_1,\ldots,y_T)$ from the problem-only prompt:
\begin{equation}
    \mathbf y \sim \pi_\theta(\cdot\mid x), \qquad
    s_t=(x,y_{<t}).
\end{equation}
At each student-visited state $s_t$, the unperturbed privileged teacher
\mzero{} has parameters $\bar\theta$ and observes both $z$ and the same student
prefix:
\begin{equation}
    p_t(v)=\pi_\theta(v\mid s_t), \qquad
    q^0_t(v)=\pi_{\bar\theta}(v\mid x,z,y_{<t}).
\end{equation}
The teacher returns a next-token distribution along the student rollout.
OPSD uses $q_t^0$ at every state and applies a pointwise forward-KL clip with
threshold $\kappa$ \citep{zhao2026selfdistilled}. \method{} retains this clip
and selects the supervision source from an expert pool.

For seed $r_j$ and radius $\sigma$, perturbation expert $j\ge1$ has parameters
\begin{equation}
    \bar\theta_j = \bar\theta + \epsilon_j,
    \qquad \epsilon_j \sim \mathcal N(0,\sigma^2 I).
    \label{eq:perturb}
\end{equation}
The noise is deterministic for each $(r_j,\sigma)$. The candidate pool includes
these perturbation experts and \mzero{}, with $\bar\theta_0=\bar\theta$.
All experts receive the same privileged information and student prefix, so
their diversity comes from the local parameter neighborhood. The teacher
snapshot and all experts remain fixed throughout training. We select a compact expert pool offline and route
one expert at each student-visited state. The radius is calibrated with
student-prefix continuations before downstream evaluation.

\subsection{Complementary Expert Selection}
\label{sec:selection}

We select experts for the corrections they add to the pool. At each reference
position, we credit a candidate only for its filtered correction beyond the
pool's current best.

Candidate selection uses teacher forcing on reference solutions. Let $y_t^*$ be
the reference token. At position $t$, the fixed problem-only model and every
candidate expert evaluate the same reference prefix:
\begin{equation}
    p^S_t(v)=\pi_{\bar\theta}(v\mid x,y^*_{<t}), \qquad
    q^j_t(v)=\pi_{\bar\theta_j}(v\mid x,z,y^*_{<t}).
\end{equation}
The shared prefixes make candidate corrections comparable at the same state
and reference token.
The fixed problem-only model provides the same baseline for every candidate,
including \mzero{}.

To focus selection on lower-confidence reference positions, we apply a
student-confidence gate (SCGate):
\begin{equation}
    g^{\mathrm{ref}}_t
    =\mathbb I[p^S_t(y_t^*)\le\tau_{\mathrm{sel}}].
\end{equation}
For each retained position, we compute the reference-token forward-KL
contribution
\begin{equation}
    c_{j,t}=q^j_t(y_t^*)
      \log\frac{q^j_t(y_t^*)}{p^S_t(y_t^*)}.
\end{equation}
We set $\kappa_{\mathrm{sel}}=\kappa$ and give no selection credit to pairs with
$c_{j,t}>\kappa_{\mathrm{sel}}$. This reference-side selection clip filters
sharp reference-token peaks that the clipped online objective may suppress.
It examines the reference token on a fixed reference prefix, while the online
clip acts on every vocabulary entry at student-visited states.

Positive teacher advantage (PTA) measures how much an expert raises the
reference-token probability. Applying both filters gives filtered PTA:
\begin{equation}
    \begin{aligned}
        A_{j,t}&=\left[q^j_t(y_t^*)-p^S_t(y_t^*)\right]_+,\\
        \widetilde A_{j,t}&=
        g^{\mathrm{ref}}_t
        \mathbb I[c_{j,t}\le\kappa_{\mathrm{sel}}] A_{j,t}.
    \end{aligned}
    \label{eq:selection_pta}
\end{equation}
A positive $\widetilde A_{j,t}$ denotes a reference-aligned, selection-feasible
correction. For a subset $S$ of candidate experts, we maximize
\begin{equation}
    \mathrm{Score}(S)=\sum_{(x,z)}\sum_t
    \max_{j\in S}\widetilde A_{j,t}.
    \label{eq:selection_score}
\end{equation}
The score rewards both newly covered positions and stronger corrections
at positions already covered. It therefore accounts for correction
magnitudes, whereas selection-feasible coverage counts positions with
positive filtered PTA.
If two experts supply similar corrections at the same positions, selecting
one reduces the other's marginal gain. An expert that improves different
positions can then become more useful to the pool.
Starting from $S=\varnothing$, with the empty-set maximum defined as zero, we
add the remaining expert that increases $\mathrm{Score}(S)$ the most.
Selection continues until $K$ experts are chosen.
In our experiments, the PTA selector does not choose \mzero{} within the first
50 greedy steps.

\subsection{MaxPeak-Anchored Quantile Routing}
\label{sec:routing}

Offline selection can reward an expert for a correction that no other expert
supplies. Choosing the anchor by pool consensus could suppress such an
individual signal. We therefore use maximum expert support to choose the
online anchor direction. At each student-visited state, expert
$j\in S$ produces $q^j_t=\pi_{\bar\theta_j}(\cdot\mid x,z,y_{<t})$.
Since a student prefix need not have an aligned reference token, we use MaxPeak
to select the highest-peak expert and take its top token as the anchor:
\begin{equation}
    j_t^*=\arg\max_j \max_{v\in\mathcal V}q^j_t(v),
    \qquad
    a_t^*=\arg\max_v q^{j_t^*}_t(v).
    \label{eq:maxpeak}
\end{equation}
This selects the token with the largest probability assigned by any expert.

We then collect experts whose top token matches the anchor:
\begin{equation}
    \mathcal E_t = \left\{j\in S:\arg\max_v q^j_t(v)=a_t^*\right\}.
    \label{eq:eligible}
\end{equation}
The MaxPeak expert $j_t^*$ belongs to this set. For each $j\in\mathcal E_t$,
the anchor probability gap is
$A^{\mathrm{on}}_{j,t}=q^j_t(a_t^*)-p_t(a_t^*)$.
At a fixed state, ranking these gaps is equivalent to ranking the experts'
anchor probabilities.
The eligible experts share a top token, but their full next-token
distributions can still differ.

A large positive gap can cause the anchor's forward-KL contribution to be
clipped, removing that term's gradient. A lower anchor probability may offer
little or no increase over the student's probability. To choose the level of
anchor support, we sort the gaps in ascending order and select the expert at
the lower discrete quantile:
\begin{equation}
    k_t=\left\lfloor q\left(|\mathcal E_t|-1\right)\right\rfloor,
    \qquad \widehat j_t=\mathrm{order}_{\mathcal E_t}(k_t).
    \label{eq:quantile}
\end{equation}
Here, $\mathrm{order}_{\mathcal E_t}$ indexes the sorted experts from zero.
The endpoints $q=0$ and $q=1$ select the smallest and largest anchor
probability gaps, respectively. Thus, $q=1$ selects a highest-peak expert.
The selected expert supplies the full-vocabulary
target $\widehat q_t=q^{\widehat j_t}_t$; the anchor token is used only for
routing. The anchor and matching set are recomputed at each state, so routing
can select different experts as the student visits new prefixes.

\subsection{Training Objective}
\label{sec:objective}

The online SCGate retains positions where the student assigns probability at
most $\tau$ to its sampled token:
$g_t=\mathbb I[p_t(y_t)\le\tau]$.
For each retained position, we apply the inherited OPSD pointwise clip to every
vocabulary-level forward-KL contribution:
\begin{equation}
    \ell_{t,v}=\widehat q_t(v)
    \log\frac{\widehat q_t(v)}{p_t(v)}, \qquad
    \overline\ell_{t,v}=\min(\ell_{t,v},\kappa).
    \label{eq:opsd_pointwise}
\end{equation}
The training loss is
\begin{equation}
    \mathcal L_{\text{\methodshort}}
    = \frac{1}{\sum_t g_t}
      \sum_t g_t \sum_{v\in\mathcal V}\overline\ell_{t,v}.
    \label{eq:objective}
\end{equation}
The denominator counts SCGate-retained positions. The pointwise clip is applied
before summing over the full vocabulary. Contributions below $\kappa$ remain
unchanged. Only the student parameters $\theta$ receive gradients.
Training repeats this process on rollouts sampled from the current student.
The pool remains fixed, but its predictions and the routed targets are
recomputed on each rollout.
Appendix~\ref{app:method_details} details the selection clip, greedy marginal
gain, and equivalent token-support form of MaxPeak.

\section{Experiments}
\label{sec:experiments}
\label{sec:analysis}
\label{sec:ablations}

\subsection{Experimental Setup}
\label{sec:experimental_setup}

\paragraph{Models and training.}
We train Qwen3-1.7B, 4B, and 8B \citep{yang2025qwen3} on 10,000 OpenThoughts
problems \citep{guha2025openthoughts} for 100 optimizer steps. At each scale,
the student and frozen experts start from the same base checkpoint.
Unless varied, \method{} uses $K=25$ experts and routing quantile $q=.75$.

\paragraph{Baselines.}
We use the Base, supervised fine-tuning (SFT), and GRPO scores from
\citet{zhao2026selfdistilled} and rerun OPSD, EOPD, and PW-OPSD
\citep{zhao2026selfdistilled,jin2026eopd,liu2026pwopsd} under the same
initialization, data, training budget, and evaluation as \method{}.
OPSD+SCGate isolates the online gate's effect with the fixed teacher \mzero{}.
All rerun baselines, including OPSD+SCGate, use full-parameter student
optimization, as does \method{}.

\paragraph{Benchmarks and metrics.}
We evaluate AIME 2024, AIME 2025, and HMMT February 2025
\citep{aime2024,aime2025,hmmt2025feb} using the OPSD sampling settings
\citep{zhao2026selfdistilled}, reporting Average@12 from 12 sampled responses
per problem. Avg. is the unweighted three-benchmark mean.
Following PW-OPSD \citep{liu2026pwopsd}, we report the step-100 checkpoint
for \method{} and all rerun baselines.
\method{}, OPSD, EOPD, and PW-OPSD use three independent runs per scale;
the 8B OPSD+SCGate control also uses three runs. We report means and sample
standard deviations for these methods. Other results are point estimates
unless noted. Appendix~\ref{app:protocol} gives all configurations and
implementation details.

\subsection{Main Results Across Model Scales}
\label{sec:main_results}

\method{} ranks first across all nine model-benchmark settings
(Table~\ref{tab:main}). Its three-benchmark average exceeds OPSD by $2.75$,
$1.67$, and $1.94$ points on 1.7B, 4B, and 8B, respectively.
Downstream ablations use Qwen3-8B and report single-run
three-benchmark Average@12 unless noted.
Appendix~\ref{app:ablation_results} gives per-benchmark ablation results,
matched single-run comparisons, and details of target compression.

\begin{table*}[t]
\centering
\caption{Main results across three Qwen3 scales with Average@12 accuracy
(\%; higher is better).}
\label{tab:main}
\begingroup
\hypersetup{hidelinks}
\small
\setlength{\tabcolsep}{2.5pt}
\renewcommand{\arraystretch}{1.16}
\definecolor{mainfirst}{HTML}{E5C88F}
\colorlet{mainsecond}{mainfirst!55!white}
\colorlet{mainthird}{mainfirst!20!white}
\newcommand{\mainresult}[2]{%
    \if\relax\detokenize{#2}\relax
        #1%
    \else
        \makebox[2.5em][r]{#1}%
        \makebox[2.3em][l]{\hspace{.15em}{\scriptsize\color{black!85}$\pm$\,#2}}%
    \fi
}
\begin{tabular*}{\textwidth}{@{\extracolsep{\fill}}ll*{6}{c}>{\columncolor{white}[\tabcolsep][0pt]}c@{}}
\toprule
\textbf{Model} & \textbf{Metric}
& \textbf{Base}$^{\dagger}$ & \textbf{SFT}$^{\dagger}$ & \textbf{GRPO}$^{\dagger}$
& \textbf{OPSD} & \textbf{EOPD} & \textbf{PW-OPSD}
& \textbf{\methodshort} \\
\midrule
\multirow{4}{*}{\shortstack[l]{\textbf{Qwen3}\\\textbf{1.7B}}} & AIME24
& \mainresult{51.50}{} & \mainresult{48.40}{} & \mainresult{51.10}{}
& \cellcolor{mainthird}\mainresult{\textit{55.46}}{0.42} & \mainresult{52.41}{0.64} & \cellcolor{mainsecond}\mainresult{\underline{57.32}}{0.70} & \cellcolor{mainfirst}\mainresult{\textbf{57.96}}{0.42} \\
 & AIME25
& \mainresult{36.70}{} & \mainresult{36.30}{} & \mainresult{38.30}{}
& \cellcolor{mainthird}\mainresult{\textit{40.93}}{0.70} & \mainresult{39.35}{0.42} & \cellcolor{mainsecond}\mainresult{\underline{41.76}}{0.70} & \cellcolor{mainfirst}\mainresult{\textbf{45.09}}{0.97} \\
 & HMMT25
& \mainresult{23.10}{} & \mainresult{22.70}{} & \mainresult{23.70}{}
& \cellcolor{mainthird}\mainresult{\textit{29.26}}{0.42} & \mainresult{25.93}{0.70} & \cellcolor{mainsecond}\mainresult{\underline{30.46}}{0.58} & \cellcolor{mainfirst}\mainresult{\textbf{30.83}}{1.11} \\
 & \textbf{Avg.}
& \mainresult{37.10}{} & \mainresult{35.80}{} & \mainresult{37.70}{}
& \cellcolor{mainthird}\mainresult{\textit{41.88}}{0.19} & \mainresult{39.23}{0.23} & \cellcolor{mainsecond}\mainresult{\underline{43.18}}{0.14} & \cellcolor{mainfirst}\mainresult{\textbf{44.63}}{0.56} \\
\midrule
\multirow{4}{*}{\shortstack[l]{\textbf{Qwen3}\\\textbf{4B}}} & AIME24
& \mainresult{74.90}{} & \mainresult{70.20}{} & \mainresult{75.60}{}
& \cellcolor{mainthird}\mainresult{\textit{76.20}}{0.16} & \mainresult{73.43}{0.58} & \cellcolor{mainsecond}\mainresult{\underline{76.48}}{0.42} & \cellcolor{mainfirst}\mainresult{\textbf{76.95}}{0.74} \\
 & AIME25
& \mainresult{66.40}{} & \mainresult{62.30}{} & \cellcolor{mainthird}\mainresult{\textit{68.10}}{}
& \mainresult{67.87}{0.70} & \mainresult{66.11}{1.00} & \cellcolor{mainsecond}\mainresult{\underline{68.70}}{0.85} & \cellcolor{mainfirst}\mainresult{\textbf{70.83}}{0.56} \\
 & HMMT25
& \mainresult{42.20}{} & \mainresult{43.40}{} & \cellcolor{mainthird}\mainresult{\textit{44.40}}{}
& \cellcolor{mainsecond}\mainresult{\underline{45.55}}{0.74} & \mainresult{42.32}{0.85} & \mainresult{42.68}{0.85} & \cellcolor{mainfirst}\mainresult{\textbf{46.85}}{0.70} \\
 & \textbf{Avg.}
& \mainresult{61.17}{} & \mainresult{58.63}{} & \cellcolor{mainthird}\mainresult{\textit{62.70}}{}
& \cellcolor{mainsecond}\mainresult{\underline{63.21}}{0.11} & \mainresult{60.62}{0.46} & \mainresult{62.62}{0.14} & \cellcolor{mainfirst}\mainresult{\textbf{64.88}}{0.39} \\
\midrule
\multirow{4}{*}{\shortstack[l]{\textbf{Qwen3}\\\textbf{8B}}} & AIME24
& \mainresult{75.80}{} & \mainresult{72.30}{} & \mainresult{76.40}{}
& \mainresult{77.41}{0.32} & \cellcolor{mainthird}\mainresult{\textit{77.50}}{0.56} & \cellcolor{mainsecond}\mainresult{\underline{77.59}}{0.64} & \cellcolor{mainfirst}\mainresult{\textbf{78.70}}{0.69} \\
 & AIME25
& \mainresult{65.60}{} & \mainresult{64.20}{} & \mainresult{68.90}{}
& \cellcolor{mainsecond}\mainresult{\underline{71.02}}{0.32} & \mainresult{70.92}{0.97} & \cellcolor{mainthird}\mainresult{\textit{70.93}}{0.70} & \cellcolor{mainfirst}\mainresult{\textbf{72.78}}{0.28} \\
 & HMMT25
& \mainresult{43.90}{} & \mainresult{42.90}{} & \cellcolor{mainthird}\mainresult{\textit{46.70}}{}
& \mainresult{45.46}{0.42} & \mainresult{45.46}{0.64} & \cellcolor{mainsecond}\mainresult{\underline{46.85}}{0.80} & \cellcolor{mainfirst}\mainresult{\textbf{48.24}}{0.89} \\
 & \textbf{Avg.}
& \mainresult{61.77}{} & \mainresult{59.80}{} & \mainresult{64.00}{}
& \cellcolor{mainthird}\mainresult{\textit{64.63}}{0.16} & \cellcolor{mainthird}\mainresult{\textit{64.63}}{0.09} & \cellcolor{mainsecond}\mainresult{\underline{65.12}}{0.33} & \cellcolor{mainfirst}\mainresult{\textbf{66.57}}{0.46} \\
\bottomrule
\end{tabular*}
\par\smallskip
\begin{minipage}{\textwidth}
\footnotesize
$^{\dagger}$: results from \citet{zhao2026selfdistilled}; other entries are our
means $\pm$ sample standard deviations over three independent training runs.
\textbf{Bold}/\underline{underlined}/\textit{italic}: first/second/third in each
row, with gold shading from darkest to lightest; ties share a rank.
Avg.: unweighted mean across the three benchmarks.
\end{minipage}
\endgroup
\end{table*}

\subsection{Complementary Expert Selection}
\label{sec:selection_analysis}

\paragraph{Selection criteria.}
We first compare expert-selection criteria with the online router and gate
held fixed (Table~\ref{tab:selection_filter_summary}). All selectors share
1,000 selection questions, 501 candidates, $\sigma=.002$, and $K=25$.
Quality Top-$K$ ranks experts by answer accuracy, following RandOpt
\citep{gan2026neuralthickets}. Greedy sample coverage adds the expert that
correctly answers the most additional questions.

Independent PTA Top-$K$ ranks experts by their total filtered PTA.
Like Quality Top-$K$, it ranks experts independently, but uses filtered
reference-token corrections as the selection criterion.
It reaches $65.65$, exceeding Quality Top-$K$ by $.65$ points.
Our PTA selector uses greedy marginal gain and reaches $66.57$, a further
$.92$-point gain. Both PTA selectors use the same per-token filtered PTA
values. They differ in whether selection accounts for corrections already
supplied by the current pool. Both comparisons show gains on all three benchmarks
(Table~\ref{tab:selector_full}).
Greedy sample coverage also rewards nonredundant contributions, but measures
them as newly solved questions. Our selector exceeds its $65.37$ average by
$1.20$ points. With routing held fixed, these results support filtered PTA
as a selection criterion and the added value of selecting complementary
corrections.

\begin{table}[!htb]
\centering
\caption{Expert selection and filtering on Qwen3-8B
(three-benchmark Average@12, \%).}
\label{tab:selection_filter_summary}
\begingroup
\small
\setlength{\tabcolsep}{1pt}
\renewcommand{\arraystretch}{1.12}
\definecolor{ablationdelta}{HTML}{855B48}
\newcommand{\ablationlabel}[1]{\begin{tabular}{@{}c@{}}#1\end{tabular}}
\newcommand{\ablationresult}[3]{%
    \begin{tabular}[t]{@{}r@{}}%
        {\normalsize #1}%
        \if\relax\detokenize{#2}\relax\else
            \hspace{.3pt}\raisebox{.65ex}{\scriptsize$\pm#2$}%
        \fi \\[-3pt]
        {\scriptsize\color{ablationdelta}%
        \if\relax\detokenize{#3}\relax\phantom{$-0.00$}\else$#3$\fi}%
    \end{tabular}%
}
\begin{tabular*}{\linewidth}{@{\extracolsep{\fill}}*{9}{c}@{}}
\toprule
\multicolumn{4}{c}{Expert selection}
& \multicolumn{2}{c}{Filtering}
& \multicolumn{2}{c}{Fixed-teacher controls}
& Full method \\
\cmidrule(lr){1-4}\cmidrule(lr){5-6}\cmidrule(lr){7-8}\cmidrule(lr){9-9}
\ablationlabel{Random\\Top-$K$}
& \ablationlabel{Quality\\Top-$K$}
& \ablationlabel{Greedy\\sample\\coverage}
& \ablationlabel{Independent\\PTA Top-$K$}
& \ablationlabel{w/o\\SCGate}
& \ablationlabel{w/o\\selection\\clip}
& OPSD
& \ablationlabel{OPSD\\+SCGate}
& \textbf{Ours} \\
\midrule
\ablationresult{64.29}{.53}{-2.28}
& \ablationresult{65.00}{}{-1.57}
& \ablationresult{65.37}{}{-1.20}
& \ablationresult{65.65}{}{-0.92}
& \ablationresult{65.28}{}{-1.29}
& \ablationresult{64.26}{}{-2.31}
& \ablationresult{64.63}{.16}{-1.94}
& \ablationresult{64.91}{.09}{-1.66}
& \ablationresult{\textbf{66.57}}{.46}{} \\
\bottomrule
\end{tabular*}
\par\smallskip
\begin{minipage}{\linewidth}
\footnotesize
Small lower-right values show differences from the full method in percentage points.
Its three-run mean and matched single-run score both equal 66.57.
Superscripts give sample standard deviations over three training runs
(Random Top-$K$: three random pools); other entries are single runs.
SCGate removal applies to both stages. The online OPSD clip is retained in
both component ablations.
\end{minipage}
\endgroup
\end{table}

\paragraph{Reference-side coverage.}
On 500 training questions held out from selection, the pool covers $26.53\%$
of reference-side SCGate-retained positions at $\sigma=.002$, compared with
$15.69\%$ for \mzero{}. Coverage requires positive filtered PTA from at least
one expert. The pool has higher coverage at every tested radius
(Table~\ref{tab:radius_summary}). Coverage also increases as experts are added
(Appendix~\ref{app:k}), supporting complementarity within the selected pool.
With reference context shared, this additional coverage comes from variation
in teacher parameters.
Newly covered reference tokens at $\sigma=.002$ include common words,
mathematical symbols, numbers, and verbs such as \texttt{find} and \texttt{need}.
Appendix~\ref{app:expert_analyses} reports the full reference-side audit and
student-prefix continuation probe.

\paragraph{SCGate and the selection clip.}
\label{sec:filtering_analysis}
We also test which positions and reference-token gains to retain.
In the reference-prefix audit, the problem-only model's top-1 confidence is at
least $.99$ at $65.63\%$ of positions. Its top-1 accuracy on these positions is
$97.67\%$, motivating supervision of lower-confidence positions.
With the fixed \mzero{} teacher, SCGate raises the three-run average from
$64.63$ to $64.91$. \method{} reaches $66.57$, a further $1.66$-point gain
(Table~\ref{tab:selection_filter_summary}). The online gate thus accounts for
only part of the improvement over OPSD.

Removing both offline and online SCGate and reselecting the pool gives
$65.28$. Reselecting without the reference-side selection clip gives $64.26$,
with both SCGate stages and the online OPSD clip retained.
This drop supports filtering expert credit during offline selection even
when the training objective already clips the loss.
Matched single-run comparisons in Appendix~\ref{app:ablation_results} give
the same ordering. The default $\tau=.99$ gives
the highest Average@12 among the four tested thresholds;
Appendix~\ref{app:scgate} reports the complete confidence audit and threshold
sensitivity results.

\FloatBarrier
\subsection{Routing on Student Prefixes}
\label{sec:transfer}
\label{sec:routing_analysis}

Offline selection uses reference prefixes, while training visits prefixes
sampled by the student. We first test the pool on student-prefix
continuations, then examine how to choose its training targets.

\paragraph{Student-prefix continuations.}
MaxPeak reaches $95.58\%$ reference-side anchor accuracy at $\sigma=.002$,
versus $91.33\%$ for \mzero{} on the same SCGate-retained positions.
We test transfer using prefixes from the frozen base student on 300 distinct
questions. These come from 150 completed wrong responses and 150 unfinished
responses. Raw MaxPeak selects the
highest-peak expert at each decoding step without quantile selection.
Both Raw MaxPeak and \mzero{} use the same prefixes and privileged references,
with a budget of 4096 new tokens. Raw MaxPeak raises continuation accuracy
from $83.33\%$ to $86.67\%$ and completion from $90.00\%$ to $94.67\%$.

Table~\ref{tab:radius_summary} connects these continuation results to
downstream accuracy across radii.
Both continuation accuracy and downstream Average@12 peak at $\sigma=.002$,
although reference-side coverage and anchor accuracy continue to rise at larger
radii. Higher reference-side coverage therefore need not yield better student
supervision. This supports calibrating the radius on student prefixes.

\begin{table}[tb]
\centering
\caption{Reference-side corrections, student-prefix continuations, and
downstream accuracy across perturbation radii on Qwen3-8B (\%).}
\label{tab:radius_summary}
\begingroup
\small
\setlength{\tabcolsep}{4pt}
\renewcommand{\arraystretch}{1.10}
\begin{tabular*}{\linewidth}{@{\extracolsep{\fill}}lrrrrr@{}}
\toprule
\multirow{2}{*}{Setting}
& \multicolumn{2}{c}{Reference prefixes}
& \multicolumn{2}{c}{Student-prefix continuations}
& \shortstack{Distilled\\student} \\
\cmidrule(lr){2-3}\cmidrule(lr){4-5}\cmidrule(l){6-6}
& SF coverage & Top-1 acc. & Accuracy & Completion & Avg. \\
\midrule
\mzero{} & 15.69 & 91.33 & 83.33 & 90.00 & n/a \\
\midrule
$\sigma=.001$ & 21.86 & 94.20 & 81.33 & 88.67 & 65.83 \\
$\sigma=.002$ & 26.53 & 95.58 & \textbf{86.67} & \textbf{94.67} & \textbf{66.57} \\
$\sigma=.004$ & 29.02 & 96.53 & 62.33 & 79.00 & 64.54 \\
$\sigma=.006$ & \textbf{31.39} & \textbf{97.31} & 27.33 & 39.33 & 63.24 \\
\bottomrule
\end{tabular*}
\par\vspace{3pt}
\begin{minipage}{\linewidth}\small
Reference metrics use SCGate-retained positions on 500 held-out questions.
For each radius, selection-feasible (SF) coverage uses the pool and top-1
accuracy uses the MaxPeak anchor. Continuations use the same 300 student
prefixes, privileged references, and 4096-token budget; all prefixes remain
in the denominator. Continuation results for each radius use Raw MaxPeak
without quantile selection.
Avg. uses \method{} with $K=25$, $q=.75$, and matched single training runs.
\end{minipage}
\endgroup
\end{table}

\begin{figure}[!htb]
    \centering
    \begin{minipage}[b]{0.32\linewidth}
        \centering
        \includegraphics[width=\linewidth]{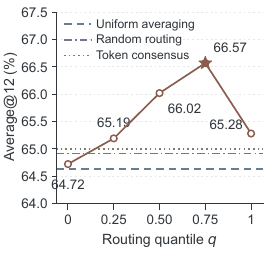}
        \par\nointerlineskip\kern1pt
        {\small (a) Quantile routing}
    \end{minipage}\hfill
    \begin{minipage}[b]{0.32\linewidth}
        \centering
        \includegraphics[width=\linewidth]{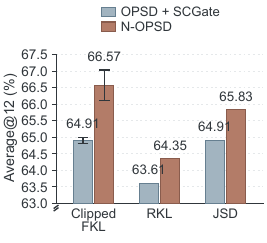}
        \par\nointerlineskip\kern1pt
        {\small (b) Online divergence}
    \end{minipage}\hfill
    \begin{minipage}[b]{0.32\linewidth}
        \centering
        \includegraphics[width=\linewidth]{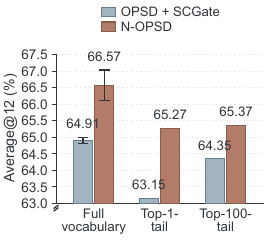}
        \par\nointerlineskip\kern1pt
        {\small (c) Target distribution}
    \end{minipage}
    \caption{Routing and distillation ablations on Qwen3-8B.
    (a) Routing rules on the same PTA-selected pool.
    (b, c) Comparisons with OPSD+SCGate under different divergences and target
    distributions, with the pool and router fixed for \method{}.
    Values are three-benchmark Average@12 (\%). Results with error bars are
    means $\pm$ sample standard deviations over three training runs;
    other results are single runs.}
    \label{fig:experiment_ablations}
\end{figure}

\paragraph{Training-time routing.}
At $\sigma=.002$, Figure~\ref{fig:experiment_ablations}(a) compares
routing rules on the same PTA-selected pool. Uniform averaging takes
the mean of all expert distributions. Token consensus equally averages the
distributions of experts voting for the most common top token.
Random routing samples one expert uniformly at each state and uses its full
distribution. All rules share the same SCGate loss normalization.
At $q=.75$, MaxPeak-anchored quantile routing gains $1.94$ points over Uniform
averaging and $1.57$ over Token consensus. Our router raises Average@12 from
$64.91$ with Random routing to $66.57$. Both routers use the same pool and supply
one expert's full distribution at each state. This gain supports choosing
that expert using the current expert predictions.

The quantile comparison fixes the pool and MaxPeak anchor rule while varying
which anchor-consistent expert supplies the full distribution. Accuracy rises
from $q=0$ to $.75$ but drops $1.29$ points at $q=1$.
Thus, the highest-peak expert need not provide the best training target,
supporting separate choices of anchor direction and support level.

\FloatBarrier
\subsection{Pool Size and Training Cost}
\label{sec:training_cost}

At $\sigma=.002$ and $q=.75$, increasing the pool from $K=1$ to $25$
raises Average@12 from $64.54$ to $66.57$ (Appendix~\ref{app:k}).
On the 500-question analysis set, $K=25$ retains $94.81\%$ of the
selection-feasible coverage at $K=50$.

Expanding to $K=50$ adds reference-side coverage, but Average@12 falls to
$64.35$ at $q=.75$. Lowering $q$ to $.50$ raises it to $65.83$, still below
the $K=25$ result. This shift in the best tested quantile supports choosing
pool size together with the level of routed support. Among the tested settings,
$K=25$ gives the highest three-benchmark average.

Using $K=25$ instead of $K=50$ reduces estimated step time by $25.63\%$,
to $1.423\times$ that of OPSD+SCGate. This estimate combines measured teacher time
with shared rollout and student computation times
(Appendix~\ref{app:timing}). Inference uses only the distilled student.

\FloatBarrier
\subsection{Robustness to Distillation Choices}
\label{sec:training_targets}

Finally, we test whether the benefit of the selected supervision persists when
the distillation objective or target distribution changes.

\paragraph{Online divergence.}
With the pool and router fixed, \method{} outperforms OPSD+SCGate under
both Jensen-Shannon divergence (JSD) and reverse KL (RKL)
(Figure~\ref{fig:experiment_ablations}(b)), with gains on all three benchmarks.
The source of supervision therefore matters beyond the default forward-KL
objective.
The clipped forward-KL reference has the highest average for \method{},
with a three-run mean of $66.57$.

\paragraph{Target distribution.}
Top-$N$-tail retains the teacher's $N$ most probable tokens and combines
the rest into one tail category. The student uses the same categories.
For Top-1-tail, supervision is a binary soft target over the top token
and the remaining mass.
We compress the distributions after routing and then apply forward-KL clipping.
\method{} retains an average advantage for both Top-1-tail and
Top-100-tail (Figure~\ref{fig:experiment_ablations}(c)), with a $2.12$-point
gain over OPSD+SCGate even for the binary target. Its full-vocabulary target
scores $1.30$ and $1.20$ points higher than these two compressed variants,
respectively.

\FloatBarrier
\section{Related Work}
\label{sec:related}

\subsection{On-Policy and Privileged Self-Distillation}

On-policy distillation supervises student-generated prefixes to reduce the
state mismatch of off-policy imitation
\citep{ross2011dagger,agarwal2024onpolicy,gu2024minillm,ko2024distillm}.
Self-distillation can condition the teacher on demonstrations, feedback, or
guiding context while training on student rollouts
\citep{shenfeld2026sdft,hubotter2026sdpo,ye2026opcd}.
OPSD conditions the teacher on a reference solution unavailable to the student
\citep{zhao2026selfdistilled}.
\citet{tian2026privilegedcontext} study the choice of privileged context,
including context selection per trajectory.
\method{} keeps the reference context shared across experts and varies
the teacher parameters.

\subsection{Adapting Supervision from a Fixed Teacher}

Fixed-teacher methods adapt the target, divergence, or loss weights.
TAID gradually interpolates student and teacher logits \citep{shing2025taid},
and DistiLLM-2 uses different divergences for teacher- and student-generated responses
\citep{ko2025distillm2}.
EOPD adapts divergence to teacher entropy \citep{jin2026eopd}, while PW-OPSD
and IW-OPD reweight token losses \citep{liu2026pwopsd,xie2026iwopd}.
Other methods use lookahead feedback, trajectory weighting, hidden-state
transitions, or purified targets
\citep{liu2026teacherhelp,lin2026renio,li2026phf,shen2026purified}.
Cal-OPD calibrates token-level discrepancies using contrasting teacher contexts
\citep{he2026calopd}.
\method{} expands the supervision source through complementary local experts.
Its routed targets can be used with divergence adaptation or loss weighting.

\subsection{Multiple Supervision Sources and Perturbation Experts}

Multi-teacher distillation includes instance-dependent teacher selection
\citep{yuan2021reinforced} and learning a distribution to sample one teacher
per update \citep{ding2024dynakd}.
For on-policy distillation, MOPD and PROMPTSD assign examples to domain- or
task-specific teachers \citep{ma2026mopd,ma2026promptsd}.
H-OPD mixes vision-language and text-only teacher distributions
\citep{yin2026hopd}, while TrustMOPD weights specialist supervision at each
student prefix \citep{sun2026trustmopd}.
Neural Thickets constructs local perturbation experts. RandOpt selects them
by task performance and uses majority voting \citep{gan2026neuralthickets}.
Its distillation variant uses SFT on generated reasoning traces and answers.
\method{} selects experts by marginal filtered PTA on reference prefixes
and routes one expert's full next-token distribution at each student-visited
state.

\section{Conclusion}

We showed that a privileged teacher's reference-aligned supervision extends
beyond its unperturbed parameter setting. \method{} turns complementary
corrections in this neighborhood into state-wise supervision for a single student.
The selected pool achieves higher reference-side selection-feasible coverage
than \mzero{}. Student-prefix continuations support using the pool
beyond reference trajectories. Matched ablations support the roles of
complementary expert selection and state-wise routing in the downstream gains.
The highest-peak expert need not provide the best training target, supporting
separate choices of anchor direction and support level.
Across Qwen3-1.7B, 4B, and 8B, \method{} improves the three-benchmark Average@12
over the original OPSD by $2.75$, $1.67$, and $1.94$ points.
Inference uses only the distilled student.
Appendix~\ref{app:limitations} discusses limitations and future work.

\bibliography{references}

\begin{thebibliography}{33}
\providecommand{\natexlab}[1]{#1}
\providecommand{\url}[1]{\texttt{#1}}
\expandafter\ifx\csname urlstyle\endcsname\relax
  \providecommand{\doi}[1]{doi: #1}\else
  \providecommand{\doi}{doi: \begingroup \urlstyle{rm}\Url}\fi

\bibitem[Agarwal et~al.(2024)Agarwal, Vieillard, Zhou, Stanczyk, Ramos~Garea,
  Geist, and Bachem]{agarwal2024onpolicy}
Rishabh Agarwal, Nino Vieillard, Yongchao Zhou, Piotr Stanczyk, Sabela
  Ramos~Garea, Matthieu Geist, and Olivier Bachem.
\newblock On-policy distillation of language models: Learning from
  self-generated mistakes.
\newblock In \emph{International Conference on Learning Representations}, 2024.

\bibitem[Cho \& Hariharan(2019)Cho and Hariharan]{cho2019efficacy}
Jang~Hyun Cho and Bharath Hariharan.
\newblock On the efficacy of knowledge distillation.
\newblock In \emph{Proceedings of the IEEE/CVF International Conference on
  Computer Vision (ICCV)}, October 2019.

\bibitem[Ding et~al.(2024)Ding, Jiang, Zhang, Guo, and Lin]{ding2024dynakd}
Zixiang Ding, Guoqing Jiang, Shuai Zhang, Lin Guo, and Wei Lin.
\newblock How to trade off the quantity and capacity of teacher ensemble:
  Learning categorical distribution to stochastically employ a teacher for
  distillation.
\newblock \emph{Proceedings of the AAAI Conference on Artificial Intelligence},
  38\penalty0 (16):\penalty0 17915--17923, 2024.
\newblock \doi{10.1609/aaai.v38i16.29746}.
\newblock URL \url{https://doi.org/10.1609/aaai.v38i16.29746}.

\bibitem[Gan \& Isola(2026)Gan and Isola]{gan2026neuralthickets}
Yulu Gan and Phillip Isola.
\newblock Neural thickets: Diverse task experts are dense around pretrained
  weights.
\newblock \emph{arXiv preprint arXiv:2603.12228}, 2026.

\bibitem[Gu et~al.(2024)Gu, Dong, Wei, and Huang]{gu2024minillm}
Yuxian Gu, Li~Dong, Furu Wei, and Minlie Huang.
\newblock {MiniLLM}: Knowledge distillation of large language models.
\newblock In \emph{International Conference on Learning Representations}, 2024.

\bibitem[Guha et~al.(2025)Guha, Marten, Keh, Raoof, Smyrnis, Bansal, Nezhurina,
  Mercat, Vu, Sprague, et~al.]{guha2025openthoughts}
Etash Guha, Ryan Marten, Sedrick Keh, Negin Raoof, Georgios Smyrnis, Hritik
  Bansal, Marianna Nezhurina, Jean Mercat, Trung Vu, Zayne Sprague, et~al.
\newblock {OpenThoughts}: Data recipes for reasoning models.
\newblock \emph{arXiv preprint arXiv:2506.04178}, 2025.

\bibitem[{Harvard-MIT Mathematics Tournament}(2025)]{hmmt2025feb}
{Harvard-MIT Mathematics Tournament}.
\newblock {HMMT February 2025: Problems and Solutions}.
\newblock \url{https://www.hmmt.org/www/archive/282}, 2025.

\bibitem[He et~al.(2026)He, Li, and Chen]{he2026calopd}
Qiangqiang He, Jin Li, and MingCai Chen.
\newblock Calibrating teacher--student discrepancy for on-policy distillation.
\newblock \emph{arXiv preprint arXiv:2609.21619}, 2026.
\newblock URL \url{https://arxiv.org/abs/2609.21619}.

\bibitem[H{\"u}botter et~al.(2026)H{\"u}botter, L{\"u}beck, Behric, Baumann,
  Bagatella, Marta, Hakimi, Shenfeld, Buening, Guestrin, and
  Krause]{hubotter2026sdpo}
Jonas H{\"u}botter, Frederike L{\"u}beck, Lejs Behric, Anton Baumann, Marco
  Bagatella, Daniel Marta, Ido Hakimi, Idan Shenfeld, Thomas~Kleine Buening,
  Carlos Guestrin, and Andreas Krause.
\newblock Reinforcement learning via self-distillation.
\newblock \emph{arXiv preprint arXiv:2601.20802}, 2026.
\newblock URL \url{https://arxiv.org/abs/2601.20802}.

\bibitem[Jin et~al.(2026)Jin, Min, Yang, Wei, Zhou, Kadhe, Baracaldo, and
  Lee]{jin2026eopd}
Woogyeol Jin, Taywon Min, Yongjin Yang, Dennis Wei, Yi~Zhou, Swanand~Ravindra
  Kadhe, Nathalie Baracaldo, and Kimin Lee.
\newblock Entropy-aware on-policy distillation of language models.
\newblock \emph{arXiv preprint arXiv:2603.07079}, 2026.

\bibitem[Ko et~al.(2024)Ko, Kim, Chen, and Yun]{ko2024distillm}
Jongwoo Ko, Sungnyun Kim, Tianyi Chen, and Se-Young Yun.
\newblock {DistiLLM}: Towards streamlined distillation for large language
  models.
\newblock In \emph{International Conference on Machine Learning}, 2024.

\bibitem[Ko et~al.(2025)Ko, Chen, Kim, Ding, Liang, Zharkov, and
  Yun]{ko2025distillm2}
Jongwoo Ko, Tianyi Chen, Sungnyun Kim, Tianyu Ding, Luming Liang, Ilya Zharkov,
  and Se-Young Yun.
\newblock {D}isti{LLM}-2: A contrastive approach boosts the distillation of
  {LLM}s.
\newblock In \emph{Proceedings of the 42nd International Conference on Machine
  Learning}, volume 267, pp.\  31044--31062, 2025.
\newblock URL \url{https://proceedings.mlr.press/v267/ko25a.html}.

\bibitem[Li et~al.(2026)Li, Zhang, Shen, and Sun]{li2026phf}
Yuhan Li, Mingxu Zhang, Dazhong Shen, and Ying Sun.
\newblock {PHF}: Privileged hidden flow for on-policy self-distillation.
\newblock \emph{arXiv preprint arXiv:2606.29340}, 2026.

\bibitem[Lin et~al.(2026)Lin, Chen, and Zhang]{lin2026renio}
Chen Lin, Kedi Chen, and Wei Zhang.
\newblock {ReNIO}: Reweighting negative trajectory importance for {LLM}
  on-policy distillation.
\newblock \emph{arXiv preprint arXiv:2606.23104}, 2026.

\bibitem[Liu et~al.(2026{\natexlab{a}})Liu, Wang, Ma, Zhang, and
  Xiao]{liu2026pwopsd}
Xiaogeng Liu, Xinyan Wang, Yingzi Ma, Yechao Zhang, and Chaowei Xiao.
\newblock When are teacher tokens reliable? position-weighted on-policy
  self-distillation for reasoning.
\newblock \emph{arXiv preprint arXiv:2605.21606}, 2026{\natexlab{a}}.

\bibitem[Liu et~al.(2026{\natexlab{b}})Liu, Lou, Guan, Ji, Lin, He, Han, Sun,
  Yu, and Lu]{liu2026teacherhelp}
Yanjiang Liu, Jie Lou, Xinyan Guan, Yuqiu Ji, Hongyu Lin, Ben He, Xianpei Han,
  Le~Sun, Xing Yu, and Yaojie Lu.
\newblock Your teacher can't help you here: Combating supervision fidelity
  decay in on-policy distillation.
\newblock \emph{arXiv preprint arXiv:2605.30833}, 2026{\natexlab{b}}.

\bibitem[Lopez-Paz et~al.(2016)Lopez-Paz, Bottou, Sch{\"o}lkopf, and
  Vapnik]{lopezpaz2016privileged}
David Lopez-Paz, L{\'e}on Bottou, Bernhard Sch{\"o}lkopf, and Vladimir Vapnik.
\newblock Unifying distillation and privileged information.
\newblock In \emph{International Conference on Learning Representations}, 2016.

\bibitem[Ma et~al.(2026{\natexlab{a}})Ma, Wei, Zhao, Zhang, Xiao, Li, Yang,
  Gao, Wang, Li, Dong, Sui, and Luo]{ma2026mopd}
Wenhan Ma, Jianyu Wei, Liang Zhao, Hailin Zhang, Bangjun Xiao, Lei Li, Qibin
  Yang, Bofei Gao, Yudong Wang, Rang Li, Jinhao Dong, Zhifang Sui, and Fuli
  Luo.
\newblock {MOPD}: Multi-teacher on-policy distillation for capability
  integration in {LLM} post-training.
\newblock \emph{arXiv preprint arXiv:2606.30406}, 2026{\natexlab{a}}.

\bibitem[Ma et~al.(2026{\natexlab{b}})Ma, Zhu, Jiang, and Xiao]{ma2026promptsd}
Yingzi Ma, Zichen Zhu, Ming Jiang, and Chaowei Xiao.
\newblock One student, many teachers: Multi-task on-policy distillation via
  soft-prompt privileged context.
\newblock \emph{arXiv preprint arXiv:2607.18293}, 2026{\natexlab{b}}.

\bibitem[{Mathematical Association of America}(2024)]{aime2024}
{Mathematical Association of America}.
\newblock {2024 American Invitational Mathematics Examination}.
\newblock \url{https://maa.org/maa-invitational-competitions/}, 2024.

\bibitem[{Mathematical Association of America}(2025)]{aime2025}
{Mathematical Association of America}.
\newblock {2025 American Invitational Mathematics Examination}.
\newblock \url{https://maa.org/maa-invitational-competitions/}, 2025.

\bibitem[Ross et~al.(2011)Ross, Gordon, and Bagnell]{ross2011dagger}
St{\'e}phane Ross, Geoffrey~J. Gordon, and J.~Andrew Bagnell.
\newblock A reduction of imitation learning and structured prediction to
  no-regret online learning.
\newblock In \emph{Proceedings of the International Conference on Artificial
  Intelligence and Statistics}, 2011.

\bibitem[Shen et~al.(2026)Shen, Tong, Yan, Shen, Chen, Ye, Hu, Miao, Wang,
  Zhao, Chen, and Ye]{shen2026purified}
Zhanming Shen, Jintao Tong, Shaotian Yan, Chen Shen, Hao Chen, Wentao Ye,
  Xiaomeng Hu, Rui Miao, Haobo Wang, Junbo Zhao, Gang Chen, and Jieping Ye.
\newblock Purified {OPSD}: On-policy self-distillation without losing how to
  think.
\newblock \emph{arXiv preprint arXiv:2607.02234}, 2026.

\bibitem[Shenfeld et~al.(2026)Shenfeld, Damani, H{\"u}botter, and
  Agrawal]{shenfeld2026sdft}
Idan Shenfeld, Mehul Damani, Jonas H{\"u}botter, and Pulkit Agrawal.
\newblock Self-distillation enables continual learning.
\newblock \emph{arXiv preprint arXiv:2601.19897}, 2026.
\newblock URL \url{https://arxiv.org/abs/2601.19897}.

\bibitem[Shing et~al.(2025)Shing, Misaki, Bao, Yokoi, and Akiba]{shing2025taid}
Makoto Shing, Kou Misaki, Han Bao, Sho Yokoi, and Takuya Akiba.
\newblock {TAID}: Temporally adaptive interpolated distillation for efficient
  knowledge transfer in language models.
\newblock In \emph{International Conference on Learning Representations}, 2025.

\bibitem[Sun et~al.(2026)Sun, Zheng, Song, Liu, Li, Jiang, Cheng, Feng, Cai,
  Fang, and Wang]{sun2026trustmopd}
Jie Sun, Mao Zheng, Mingyang Song, Zeyuan Liu, Gengsheng Li, Houcheng Jiang,
  Yilin Cheng, Bichuan Feng, Yuchen Cai, Junfeng Fang, and Xiang Wang.
\newblock Distill what you trust: Reliability-aware multi-teacher on-policy
  distillation.
\newblock \emph{arXiv preprint arXiv:2609.23697}, 2026.
\newblock URL \url{https://arxiv.org/abs/2609.23697}.

\bibitem[Tian et~al.(2026)Tian, Liu, Jiang, Dong, Li, Ding, Guo, Li, Hou, Wang,
  and Wang]{tian2026privilegedcontext}
Kanghui Tian, Siyuan Liu, Tianxiang Jiang, Shuai Dong, Yizhuo Li, Tian Ding,
  Yuan Guo, Songze Li, Haowen Hou, Congcong Wang, and Yi~Wang.
\newblock What should a self-teacher see? privileged context design for
  on-policy self-distillation.
\newblock \emph{arXiv preprint arXiv:2609.25623}, 2026.
\newblock URL \url{https://arxiv.org/abs/2609.25623}.

\bibitem[Xie et~al.(2026)Xie, Zhu, Wen, Chen, and Wang]{xie2026iwopd}
Yan Xie, Sijie Zhu, Tiansheng Wen, Bo~Chen, and Yifei Wang.
\newblock On the position bias of on-policy distillation.
\newblock \emph{arXiv preprint arXiv:2606.22600}, 2026.

\bibitem[Yang et~al.(2025)Yang, Li, Yang, Zhang, Hui, Zheng, Yu, Gao, Huang,
  Lv, et~al.]{yang2025qwen3}
An~Yang, Anfeng Li, Baosong Yang, Beichen Zhang, Binyuan Hui, Bo~Zheng, Bowen
  Yu, Chang Gao, Chengen Huang, Chenxu Lv, et~al.
\newblock {Qwen3} technical report.
\newblock \emph{arXiv preprint arXiv:2505.09388}, 2025.

\bibitem[Ye et~al.(2026)Ye, Dong, Wu, Huang, and Wei]{ye2026opcd}
Tianzhu Ye, Li~Dong, Xun Wu, Shaohan Huang, and Furu Wei.
\newblock On-policy context distillation for language models.
\newblock \emph{arXiv preprint arXiv:2602.12275}, 2026.
\newblock URL \url{https://arxiv.org/abs/2602.12275}.

\bibitem[Yin et~al.(2026)Yin, Yao, Cai, Chen, Wang, Yang, Su, and
  Zhao]{yin2026hopd}
Qixiang Yin, Huanjin Yao, Yuchen Cai, Jianghao Chen, Ziyi Wang, Min Yang, Fei
  Su, and Zhicheng Zhao.
\newblock {H-OPD}: Confidence aware heterogeneous multi-teacher multimodal
  on-policy distillation.
\newblock \emph{arXiv preprint arXiv:2607.02592}, 2026.

\bibitem[Yuan et~al.(2021)Yuan, Shou, Pei, Lin, Gong, Fu, and
  Jiang]{yuan2021reinforced}
Fei Yuan, Linjun Shou, Jian Pei, Wutao Lin, Ming Gong, Yan Fu, and Daxin Jiang.
\newblock Reinforced multi-teacher selection for knowledge distillation.
\newblock \emph{Proceedings of the AAAI Conference on Artificial Intelligence},
  35\penalty0 (16):\penalty0 14284--14291, 2021.
\newblock \doi{10.1609/aaai.v35i16.17680}.
\newblock URL \url{https://doi.org/10.1609/aaai.v35i16.17680}.

\bibitem[Zhao et~al.(2026)Zhao, Xie, Liu, Huang, Pang, Chen, and
  Grover]{zhao2026selfdistilled}
Siyan Zhao, Zhihui Xie, Mengchen Liu, Jing Huang, Guan Pang, Feiyu Chen, and
  Aditya Grover.
\newblock Self-distilled reasoner: On-policy self-distillation for large
  language models.
\newblock \emph{arXiv preprint arXiv:2601.18734}, 2026.

\end{thebibliography}
\bibliographystyle{references}

\appendix

\newcommand{\appendixtablestyle}{%
    \small
    \setlength{\tabcolsep}{4pt}%
    \renewcommand{\arraystretch}{1.10}%
}
\newcommand{\appendixtablenote}[1]{%
    \par\vspace{3pt}%
    \begin{minipage}{\linewidth}\small #1\end{minipage}%
}
\section{Method Details}
\label{app:method_details}

\subsection{Reference-Side Selection Clip}

The OPSD pointwise clip in Eq.~\ref{eq:opsd_pointwise} acts on each
vocabulary-level forward-KL contribution before summation. When
$\ell_{t,v}>\kappa$, the clipped value is constant and contributes no gradient.
Contributions below the threshold remain unchanged.

The reference-side selection clip gives no credit to candidate-position pairs
with $c_{j,t}>\kappa_{\mathrm{sel}}$. It prevents experts from gaining
selection credit through sharp reference-token peaks that the clipped online
objective may suppress. This is a selection proxy motivated by the OPSD
objective. It examines one reference-token contribution under a fixed reference
prefix. The online clip instead examines every vocabulary-level contribution
from the routed expert at student-visited states. Passing the selection clip
does not determine whether an online contribution will be clipped.

\subsection{Greedy Marginal Gain}

For the selection objective in Eq.~\ref{eq:selection_score}, the gain from
adding expert $j$ to the current pool $S$ is
\begin{equation}
    \Delta(j\mid S)=\sum_{(x,z)}\sum_t
    \left[\widetilde A_{j,t}-
    \max_{i\in S}\widetilde A_{i,t}\right]_+.
\end{equation}
Each step adds the remaining expert with the largest $\Delta(j\mid S)$.
We rank all 501 candidates in greedy insertion order and retain the first $K$.
The coverage gains on the analysis set in Appendix~\ref{app:k} motivate $K=25$.

\subsection{MaxPeak Equivalence}
\label{app:maxpeak_equivalence}

At a student-visited state, each token's maximum expert support is
\begin{equation}
    u_t(v)=\max_{j\in S}q^j_t(v),
    \qquad
    a_t^*\in\arg\max_{v\in\mathcal V}u_t(v).
    \label{eq:pool_support}
\end{equation}
The two maxima can be exchanged:
\begin{equation}
    \max_{v\in\mathcal V}u_t(v)
    =\max_{v\in\mathcal V}\max_{j\in S}q^j_t(v)
    =\max_{j\in S}\max_{v\in\mathcal V}q^j_t(v).
    \label{eq:maxpeak_equivalence}
\end{equation}
Selecting the highest-peak expert and then its top token therefore gives a
maximizer of $u_t$, as used by MaxPeak in Eq.~\ref{eq:maxpeak}.
This online score uses expert probabilities on the current student prefix.
Offline selection uses filtered reference-token PTA.

\section{Complete Experimental Protocol}
\label{app:protocol}

\subsection{Expert Construction and Selection}

\paragraph{Selection and audit data.}
For Qwen3-8B, we randomly sample two disjoint question sets from the OPSD
training data. The selection set contains 1,000 questions and the analysis set
contains 500. This matches expert selection and analysis to the student
training distribution. Their identifiers are
\texttt{openthoughts\_train\_selection\_1000} ($\mathcal D_{\mathrm{sel}}$) and
\texttt{openthoughts\_test\_indep\_500}, respectively. The analysis set and all
three downstream benchmarks are excluded from expert selection.

\paragraph{Candidates and thresholds.}
For each model scale, we sample 500 perturbation seeds from the scale's own
frozen base checkpoint. Together with \mzero{}, the resulting experts form a
pool of 501 candidates. Expert selection uses a split disjoint from AIME and
HMMT. The 8B radius grid uses
$\sigma\in\{.001,.002,.004,.006\}$, $\tau_{\mathrm{sel}}=.99$, and
$\kappa_{\mathrm{sel}}=.06$. The first 25 experts in greedy insertion order are
retained. The reference-side selection clip excludes candidate-position
pairs whose reference-token FKL contribution exceeds the threshold on fixed
reference prefixes. The OPSD pointwise clip acts separately on each
vocabulary-level forward-KL contribution from the routed expert distribution at
student-visited states. The calibrated perturbation radii are $\sigma=.0006$,
$.0012$, and $.002$ for Qwen3-1.7B, 4B, and 8B.

\subsection{Selector Definitions}
\label{app:selectors}

Quality Top-$K$ and Greedy sample coverage evaluate one generated response per
expert and question. Let $N=|\mathcal D_{\mathrm{sel}}|$. For candidate expert
$j$ and question $i$, define
\begin{equation}
    \begin{aligned}
        b_{j,i}&=\mathbb I[\text{expert }j\text{ answers question }i\text{ correctly}],\\
        Q_j&=\frac{1}{N}\sum_{i=1}^{N}b_{j,i}.
    \end{aligned}
    \label{eq:selector_quality}
\end{equation}
Both selection scores use final-answer correctness without reference-side
SCGate or the selection clip.

\paragraph{Quality Top-$K$.}
We select the $K$ experts with the highest $Q_j$. Ties follow the fixed
candidate order.

\paragraph{Greedy sample coverage.}
Let $C_j=\{i:b_{j,i}=1\}$ be the questions answered correctly by expert $j$.
The selection objective counts questions answered correctly by at least one
selected expert:
\begin{equation}
    F(S)=\left|\bigcup_{j\in S}C_j\right|.
    \label{eq:sample_coverage}
\end{equation}
Starting from $S=\varnothing$, each step adds the expert with the largest
increase in coverage:
\begin{equation}
    j^\star=\arg\max_{j\notin S}
    \left|C_j\setminus\bigcup_{k\in S}C_k\right|,
    \qquad S\leftarrow S\cup\{j^\star\}.
    \label{eq:sample_coverage_greedy}
\end{equation}
We recompute the gains after each addition. Ties favor higher $Q_j$, then the
fixed candidate order. Selection continues until $K$ experts are chosen.
Once coverage saturates, this tie rule fills the pool by $Q_j$.

\paragraph{Independent PTA Top-$K$.}
We score each expert by its total filtered PTA on the selection set:
\begin{equation}
    R_j=\sum_{(x,z)\in\mathcal D_{\mathrm{sel}}}\sum_t
    \widetilde A_{j,t}.
    \label{eq:independent_pta}
\end{equation}
We rank experts once by $R_j$ and select the top $K$.
This rule uses the same reference-side SCGate and selection clip as the PTA
selector. Scores remain fixed as experts are added to the pool.

\subsection{Token Consensus Routing}
\label{app:consensus}

Token consensus constructs the training target from the pool $S$ of 25 experts
selected by the PTA selector. At each student prefix, every expert casts one
vote for its top-1 token:
\begin{equation}
    a^j_t=\arg\max_{v\in\mathcal V}q^j_t(v),\qquad
    n_t(v)=\sum_{j\in S}\mathbb I[a^j_t=v].
\end{equation}
The consensus token is
\begin{equation}
    a_t^{\mathrm{cons}}\in\arg\max_{v\in\mathcal V}n_t(v).
\end{equation}
A token can be selected without receiving more than half of the votes.
Ties favor the largest
$\sum_{j\in S:a^j_t=v}q^j_t(v)$, then the fixed expert order.
We collect all experts that support the consensus token:
\begin{equation}
    \mathcal E_t^{\mathrm{cons}}=
    \left\{j\in S:a^j_t=a_t^{\mathrm{cons}}\right\}.
    \label{eq:consensus_experts}
\end{equation}
We average their full vocabulary distributions with equal weights:
\begin{equation}
    \widehat q_t^{\mathrm{cons}}(v)=
    \frac{1}{|\mathcal E_t^{\mathrm{cons}}|}
    \sum_{j\in\mathcal E_t^{\mathrm{cons}}}q^j_t(v).
    \label{eq:consensus_target}
\end{equation}
As in MaxPeak-anchored quantile routing, the loss denominator is $\sum_t g_t$
(Eq.~\ref{eq:objective}).

\subsection{Training Configuration}

We use full-parameter student optimization for \method{} and all rerun baselines.
All three model scales use learning rate $10^{-6}$ and effective batch size 64.
Training lasts 100 optimizer steps with checkpoints saved every 20 steps.
The 8B experiments use eight GPUs, per-device batch size 4, and a gradient
accumulation factor of 2. For \method{}, we use forward KL
($\beta=0$ in the generalized divergence implementation) and a
full-vocabulary loss. Unless varied, the defaults are $K=25$, $q=.75$,
online SCGate threshold $\tau=.99$, and
$\kappa_{\mathrm{sel}}=\kappa=.06$. All expert parameters remain fixed, and
only the student is updated. Token-level post-sum clipping is disabled.
Gradient checkpointing and bfloat16 are enabled.

\paragraph{Memory and computation.}
For each expert forward pass, we add its deterministic perturbation, compute
the distribution, and restore the base weights.
Materializing all $K$ expert distributions would require
$O(KBS|\mathcal V|)$ memory. We instead store $O(KBS)$ routing metadata on CPU
in a first streaming pass and materialize one expert distribution at a time in
a second pass. Table~\ref{tab:k} reports estimated step time relative to OPSD+SCGate.
Appendix~\ref{app:timing} gives the timing measurements and calculation.
Inference uses only the distilled student.

\paragraph{Rollouts and filtering.}
Student rollouts use at most 1024 tokens, temperature 1.1, top-$p=.95$, and
top-$k=20$. Thinking is disabled in the student prompt and enabled in the
privileged prompt. The student sees the problem alone, and the teacher receives
the same completion with the privileged reference context. In \method{},
SCGate filters a position when the student probability of its sampled token
exceeds the threshold.
The OPSD+SCGate control applies the same online gate to the fixed \mzero{}
target and has no offline expert-selection stage. In the SCGate ablation, we
remove both the reference-side and online gates and reselect the expert pool.
In the selection-clip ablation, we reselect the pool without the reference-side
selection clip while retaining the online OPSD pointwise clip.

\subsection{Baseline Implementations}
\label{app:baselines}

OPSD, EOPD, and PW-OPSD use the frozen reference-conditioned teacher
\mzero{}. The student generates a rollout from the problem alone, and the
teacher evaluates that rollout with access to the reference solution and final
answer. OPSD applies the clipped forward-KL contribution in
Eq.~\ref{eq:opsd_pointwise} to $q_t^0$ at every rollout position, with uniform
position weights.

For EOPD, we apply the entropy-aware distillation objective of
\citet{jin2026eopd} using this privileged teacher. This comparison tests the
objective in the self-distillation setting, with the reference-conditioned
\mzero{} supplying the teacher distributions. PW-OPSD uses the
position-dependent loss weighting proposed by \citet{liu2026pwopsd}, using
the same rollout protocol and reference-conditioned teacher.

\subsection{Evaluation and Statistics}

Our runs use thinking mode, temperature 1.0, top-$p=.95$, top-$k=-1$,
maximum 38,912 new tokens, and \texttt{val.n=12}. Following PW-OPSD
\citep{liu2026pwopsd}, we report the step-100 checkpoint for \method{},
all rerun baselines, and all ablation variants.
Base, SFT, and GRPO results are taken from
\citet{zhao2026selfdistilled} under the source paper's evaluation and
checkpoint-selection protocols. The other baseline results are our reruns.
We conduct three independent training runs for \method{}, OPSD, EOPD, and
PW-OPSD at each scale. The Qwen3-8B OPSD+SCGate control also uses three
independent training runs. We report the mean and sample standard deviation
across runs. For Avg., we use each run's three-benchmark mean. The
remaining baselines and analysis variants are point estimates unless noted
otherwise.

\section{Complete Ablation Results}
\label{app:ablation_results}

The following tables give the results for each benchmark corresponding to the
averages in Section~\ref{sec:experiments}. All experiments use Qwen3-8B and
Average@12. Results are matched single runs unless noted otherwise.

\subsection{Expert Selection}

All selectors use the same 1,000 selection questions, 501 candidates,
$\sigma=.002$, and $K=25$. Training uses MaxPeak-anchored quantile routing
with $q=.75$ and online SCGate with $\tau=.99$.

\begin{table}[!htbp]
\centering
\caption{Expert-selection accuracy (\%) on Qwen3-8B with Average@12.}
\label{tab:selector_full}
\appendixtablestyle
\begin{tabular*}{\linewidth}{@{\extracolsep{\fill}}lrrrr@{}}
\toprule
Selector & AIME24 & AIME25 & HMMT25 & Avg. \\
\midrule
Random Top-$K$ & 77.04 $\pm$ .85 & 70.28 $\pm$ .48
& 45.56 $\pm$ .73 & 64.29 $\pm$ .53 \\
Quality Top-$K$ & 77.50 & 70.83 & 46.67 & 65.00 \\
Greedy sample coverage & 77.78 & 71.67 & 46.67 & 65.37 \\
Independent PTA Top-$K$ & 77.78 & 71.39 & 47.78 & 65.65 \\
PTA selector (Ours) & \textbf{78.33} & \textbf{72.78}
& \textbf{48.61} & \textbf{66.57} \\
\bottomrule
\end{tabular*}
\appendixtablenote{Random Top-$K$: mean $\pm$ sample standard deviation over three pools. Other rows are single runs. Both PTA selectors use the same reference-side SCGate and selection clip.}
\end{table}

\subsection{Online Routing}

We compare Uniform averaging, Random routing, Token consensus, and
MaxPeak-anchored quantile routing on the same PTA-selected pool at $K=25$ and
$\sigma=.002$. Random routing draws $\widehat j_t$ uniformly from $S$ at each
student-visited state and uses the full distribution
$\widehat q_t=q_t^{\widehat j_t}$. It retains the online SCGate and clipped
forward-KL objective. The quantile comparison keeps the MaxPeak anchor rule fixed.

\begin{table}[!htbp]
\centering
\caption{Routing accuracy (\%) at $\sigma=.002$ and $K=25$ on Qwen3-8B.}
\label{tab:routing_full}
\appendixtablestyle
\begin{tabular*}{\linewidth}{@{\extracolsep{\fill}}lcrrrr@{}}
\toprule
Routing rule & $q$ & AIME24 & AIME25 & HMMT25 & Avg. \\
\midrule
Uniform averaging & n/a & 77.78 & 69.17 & 46.94 & 64.63 \\
Random routing & n/a & \textbf{78.33} & 70.28 & 46.11 & 64.91 \\
Token consensus & n/a & 77.22 & 71.39 & 46.39 & 65.00 \\
\midrule
\multirow{5}{*}{\shortstack[l]{MaxPeak-anchored\\quantile}} & 0   & 77.22 & 70.83 & 46.11 & 64.72 \\
                 & .25 & 77.50 & 71.11 & 46.94 & 65.19 \\
                 & .50 & 78.06 & 72.22 & 47.78 & 66.02 \\
                 & .75 & \textbf{78.33} & \textbf{72.78}
                 & \textbf{48.61} & \textbf{66.57} \\
                 & 1   & 77.22 & 71.67 & 46.94 & 65.28 \\
\bottomrule
\end{tabular*}
\appendixtablenote{All scores use Average@12. MaxPeak selects the anchor token; $q$ selects the rank of its probability gap among anchor-consistent experts.}
\end{table}

\subsection{Divergence and Target Distribution}
\label{app:training_targets}

These ablations use the same 100-step training budget. We report the
step-100 checkpoint for each run. Both supervision sources use online SCGate with
$\tau=.99$. All \method{} variants retain the selected pool and router at
$\sigma=.002$, $K=25$, and $q=.75$.
Table~\ref{tab:training_targets_full} reports the complete results from single
training runs. The full-vocabulary forward-KL reference is the single run
reported in Table~\ref{tab:selection_filters_full}.

\begin{table}[!htbp]
\centering
\caption{Divergence and target-distribution accuracy (\%) on Qwen3-8B.}
\label{tab:training_targets_full}
\appendixtablestyle
\begin{tabular*}{\linewidth}{@{\extracolsep{\fill}}llrrrr@{}}
\toprule
Source & Setting & AIME24 & AIME25 & HMMT25 & Avg. \\
\midrule
OPSD + SCGate & JSD & 77.50 & 70.28 & 46.94 & 64.91 \\
\methodshort{} & JSD & 78.33 & 71.67 & 47.50 & \textbf{65.83} \\
OPSD + SCGate & RKL & 76.94 & 69.17 & 44.72 & 63.61 \\
\methodshort{} & RKL & 77.22 & 70.00 & 45.83 & \textbf{64.35} \\
\midrule
OPSD + SCGate & Top-1-tail & 74.44 & 68.06 & 46.94 & 63.15 \\
\methodshort{} & Top-1-tail & 76.94 & 71.94 & 46.94 & \textbf{65.27} \\
OPSD + SCGate & Top-100-tail & 76.39 & 72.50 & 44.17 & 64.35 \\
\methodshort{} & Top-100-tail & 77.22 & 70.83 & 48.06 & \textbf{65.37} \\
\midrule
\methodshort{} & Full-vocabulary FKL & 78.33 & 72.78 & 48.61 & 66.57 \\
\bottomrule
\end{tabular*}
\appendixtablenote{Average@12; all entries are single runs. JSD and RKL use the full vocabulary; Top-$N$-tail uses clipped forward KL. The last row repeats the full-vocabulary FKL reference. Avg. is the unweighted mean across benchmarks.}
\end{table}

\paragraph{Top-$N$-tail construction.}
Let $q_t$ denote the teacher target, either $q_t^0$ for OPSD+SCGate or
$\widehat q_t$ for \method{}. At each state, we select its $N$ most probable
tokens, $H_{N,t}=\{v_1,\ldots,v_N\}$. We use this same set to compress the
teacher and student distributions into $N+1$ categories:
\begin{equation}
    \begin{aligned}
        q^{(N)}_{t,i}&=q_t(v_i), \quad
        p^{(N)}_{t,i}=p_t(v_i), \qquad i=1,\ldots,N,\\
        q^{(N)}_{t,N+1}&=\sum_{v\notin H_{N,t}}q_t(v), \quad
        p^{(N)}_{t,N+1}=\sum_{v\notin H_{N,t}}p_t(v).
    \end{aligned}
    \label{eq:topn_tail}
\end{equation}
The final category preserves the total tail probability but does not match
the probabilities of individual tokens within the tail. Top-1-tail therefore
provides a binary soft target, while Top-100-tail has 101 categories.

\paragraph{Clipping after compression.}
We compute one forward-KL contribution per category and clip it before summing:
\begin{equation}
    \ell^{(N)}_{t,i}=q^{(N)}_{t,i}
    \log\frac{q^{(N)}_{t,i}}{p^{(N)}_{t,i}}, \qquad
    L^{(N)}_t=\sum_{i=1}^{N+1}\min(\ell^{(N)}_{t,i},\kappa),
    \quad \kappa=.06.
    \label{eq:topn_tail_clip}
\end{equation}
The training loss averages $L^{(N)}_t$ over SCGate-retained positions, using
the denominator in Eq.~\ref{eq:objective}. The aggregated tail receives one
clip, whereas the full-vocabulary objective clips each token separately.

\subsection{SCGate and the Reference-Side Selection Clip}

OPSD+SCGate applies the online gate to the fixed \mzero{} teacher.
For the component ablations, we reselect the expert pool after removing both
SCGate stages or the reference-side selection clip. The online OPSD pointwise
clip is retained in both ablations.

\begin{table}[!htbp]
\centering
\caption{SCGate control and selection-filter ablations on Qwen3-8B (Average@12, \%).}
\label{tab:selection_filters_full}
\appendixtablestyle
\begin{tabular*}{\linewidth}{@{\extracolsep{\fill}}lrrrr@{}}
\toprule
Variant & AIME24 & AIME25 & HMMT25 & Avg. \\
\midrule
\multicolumn{5}{l}{\emph{Three independent runs}} \\
OPSD & 77.41 $\pm$ .32 & 71.02 $\pm$ .32
& 45.46 $\pm$ .42 & 64.63 $\pm$ .16 \\
OPSD + SCGate & 77.87 $\pm$ .70 & 71.11 $\pm$ .48
& 45.74 $\pm$ .42 & 64.91 $\pm$ .09 \\
\methodshort{} & \textbf{78.70 $\pm$ .69} & \textbf{72.78 $\pm$ .28}
& \textbf{48.24 $\pm$ .89} & \textbf{66.57 $\pm$ .46} \\
\midrule
\multicolumn{5}{l}{\emph{Matched single-run component ablations}} \\
Full \methodshort{} & \textbf{78.33} & \textbf{72.78}
& \textbf{48.61} & \textbf{66.57} \\
w/o SCGate (both stages) & 77.78 & 71.94 & 46.11 & 65.28 \\
w/o selection clip
& 76.94 & 70.83 & 45.00 & 64.26 \\
\bottomrule
\end{tabular*}
\appendixtablenote{First block: mean $\pm$ sample standard deviation over three independent runs. Second block: matched single runs at $\sigma=.002$, $K=25$, $q=.75$. Removing SCGate disables both selection and training gates; removing the reference-side selection clip retains the online OPSD clip.}
\end{table}

SCGate improves the fixed-teacher average by $0.28$ points. In the matched
single-run comparison, removing both SCGate stages reduces the average by
approximately $1.3$ points. Removing the reference-side selection clip reduces
it by $2.31$ points.

\section{Parameter Sensitivity and Training Cost}
\label{app:sensitivity}

Downstream accuracy results in this section use Qwen3-8B, Average@12, and
matched single runs.

\subsection{Perturbation Radius}
\label{app:radius}

The perturbation radius controls how much expert behavior changes.
Small $\sigma$ keeps experts close to \mzero{} and may reveal little
complementarity. Larger $\sigma$ can expose more reference-aligned corrections,
but can also increase peak probabilities and expert-student mismatch.
We train the full method at each radius while fixing $K=25$ and $q=.75$.
Table~\ref{tab:radius} tests whether the radius chosen by the student-prefix
probe also gives the best downstream result.

\begin{table}[!htbp]
\centering
\caption{Perturbation-radius accuracy (\%) with $K=25$ and $q=.75$ (Qwen3-8B, Average@12).}
\label{tab:radius}
\appendixtablestyle
\begin{tabular*}{\linewidth}{@{\extracolsep{\fill}}crrrr@{}}
\toprule
$\sigma$ & AIME24 & AIME25 & HMMT25 & Avg. \\
\midrule
.001 & 77.78 & 72.22 & 47.50 & 65.83 \\
.002 & \textbf{78.33} & \textbf{72.78} & \textbf{48.61} & \textbf{66.57} \\
.004 & 77.78 & 70.56 & 45.28 & 64.54 \\
.006 & 76.39 & 69.72 & 43.61 & 63.24 \\
\bottomrule
\end{tabular*}
\end{table}

The three-benchmark average reaches $66.57$ at $\sigma=.002$, compared with
$65.83$, $64.54$, and $63.24$ at $.001$, $.004$, and $.006$. The $.002$
setting is also best on each benchmark. Both the continuation probe and
downstream evaluation favor $\sigma=.002$ among the tested radii.

\subsection{SCGate Calibration and Sensitivity}
\label{app:scgate}

The confidence audit uses the same 331,246 Qwen3-8B reference positions as
Appendix~\ref{app:reference}. For each threshold, we select positions where the
student's top prediction reaches that confidence and measure top-1 accuracy on
those positions. The downstream columns use the same threshold for online SCGate.

At $\tau=.99$, 217,400 reference positions ($65.63\%$) are in the
high-confidence group. The problem-only model reaches $97.67\%$ top-1 accuracy
in this group. This motivates focusing supervision on lower-confidence
positions.

\begin{table}[!htbp]
\centering
\caption{SCGate confidence audit and downstream accuracy at $\sigma=.002$, $K=25$, $q=.75$.}
\label{tab:scgate_full}
\appendixtablestyle
\begin{tabular*}{\linewidth}{@{\extracolsep{\fill}}crrrrrrr@{}}
\toprule
\multirow{2}{*}{$\tau$} & \multicolumn{3}{c}{High-confidence reference positions}
& \multicolumn{4}{c}{Downstream Average@12 (\%)} \\
\cmidrule(lr){2-4}\cmidrule(l){5-8}
& Count & Share (\%) & Acc. (\%) & AIME24 & AIME25 & HMMT25 & Avg. \\
\midrule
.95  & 242,255 & 73.13 & 96.02 & 77.78 & 71.39 & 46.67 & 65.28 \\
.98  & 227,137 & 68.57 & 97.09 & 77.50 & 72.22 & 47.50 & 65.74 \\
.99  & 217,400 & 65.63 & 97.67 & 78.33 & 72.78 & 48.61 & 66.57 \\
.995 & 207,988 & 62.79 & 98.10 & 77.78 & 71.11 & 47.22 & 65.37 \\
\bottomrule
\end{tabular*}
\end{table}

\subsection{Expert Count}
\label{app:k}

Table~\ref{tab:k} reports selection-feasible coverage, downstream accuracy,
and estimated step time as the expert pool grows. We measure coverage on the
500-question analysis set using the first $K$ experts in the greedy order
determined on the selection set. Coverage is normalized by $K=50$ and
estimated step time by OPSD+SCGate.

\begin{table}[!htbp]
\centering
\caption{Expert-count ablation at $\sigma=.002$ on Qwen3-8B (Average@12, \%).}
\label{tab:k}
\appendixtablestyle
\begin{tabular*}{\linewidth}{@{\extracolsep{\fill}}ccrrrrrr@{}}
\toprule
$K$ & $q$ & \shortstack{Rel. SF\\cov. (\%)} & \shortstack{Est. rel.\\step time}
& AIME24 & AIME25 & HMMT25 & Avg. \\
\midrule
1  & .75 & 63.47 & 1.010$\times$ & 76.94 & 70.28 & 46.39 & 64.54 \\
5  & .75 & 76.63 & 1.082$\times$ & 77.78 & 70.83 & 47.22 & 65.28 \\
10 & .75 & 84.48 & 1.193$\times$ & 77.50 & 71.67 & 47.78 & 65.65 \\
25 & .75 & 94.81 & 1.423$\times$ & 78.33 & \textbf{72.78}
& 48.61 & \textbf{66.57} \\
\multirow{3}{*}{50} & .75 & \multirow{3}{*}{100.00}
& \multirow{3}{*}{1.914$\times$} & 77.78 & 70.83 & 44.44 & 64.35 \\
& .50 & & & \textbf{78.89} & 69.44 & \textbf{49.17} & 65.83 \\
& .25 & & & 77.50 & 71.94 & 45.83 & 65.09 \\
\bottomrule
\end{tabular*}
\appendixtablenote{Coverage is normalized by $K=50$ and step time by OPSD+SCGate. Timing uses the same student trajectories (Appendix~\ref{app:timing}). Lower $q$ values at $K=50$ probe the relation between quantile and pool size (Section~\ref{sec:routing_analysis}).}
\end{table}

At $K=25$, the pool retains $94.81\%$ of the $K=50$ selection-feasible
coverage. It reaches the highest tested Average@12 of $66.57$, compared with
$65.83$ for the best tested $K=50$ setting. Its estimated step time is
$1.423\times$ that of OPSD+SCGate and $25.63\%$ lower than at $K=50$.
The best tested quantile changes from $q=.75$ at $K=25$ to $q=.50$ at $K=50$.
At $K=50$, $q=.50$ improves Average@12 by $1.48$ points over $q=.75$.

\begin{table}[!htbp]
\centering
\caption{Coverage gains on the 500-question analysis set at $\sigma=.002$.}
\label{tab:k_curve}
\appendixtablestyle
\begin{tabular*}{\linewidth}{@{\extracolsep{\fill}}lrrrrr@{}}
\toprule
Expert rank $K$ & 1 & 5 & 10 & 25 & 50 \\
\midrule
Marginal share (\%) & 63.47 & 2.18 & 1.21 & .39 & .15 \\
Cumulative share (\%) & 63.47 & 76.63 & 84.48 & 94.81 & 100.00 \\
\bottomrule
\end{tabular*}
\appendixtablenote{Marginal share measures the coverage added by the expert at that rank. Both shares are normalized by cumulative selection-feasible coverage at $K=50$.}
\end{table}

The marginal share falls from $63.47\%$ for the first expert to $.39\%$ at rank
25 and $.15\%$ at rank 50. The curve shows that most available
selection-feasible coverage is already present at $K=25$.

\subsection{Training-Step Timing}
\label{app:timing}

We measure teacher time on the same cached student trajectories. The unperturbed
teacher \mzero{} used by OPSD+SCGate takes $2.0184$ seconds in one forward pass.
For \method{}, Pass 1 computes routing metadata, and Pass 2 reruns experts to
build the target distribution. Their sum is the teacher time in
Table~\ref{tab:expert_timing}.

We estimate step time using the same rollout time ($78.215$ seconds) and
student computation time ($82.443$ seconds) for all settings. Student computation
includes the forward pass, loss computation, backward pass, and other training
operations. Adding the measured \mzero{} teacher time gives an OPSD+SCGate
baseline of $162.6764$ seconds. For \method{}, we add $0.012$ seconds for routing,
so estimated step time equals teacher time plus $160.670$ seconds. Both teacher
time and estimated step time are normalized by their OPSD+SCGate values.

\begin{table}[!htbp]
\centering
\caption{Measured teacher time and estimated training-step time on Qwen3-8B.}
\label{tab:expert_timing}
\appendixtablestyle
\begin{tabular*}{\linewidth}{@{\extracolsep{\fill}}lrrrrrr@{}}
\toprule
Setting & \shortstack{Pass 1\\(s)} & \shortstack{Pass 2\\(s)}
& \shortstack{Teacher\\time (s)} & \shortstack{Relative\\teacher time}
& \shortstack{Est. step\\time (s)} & \shortstack{Est. relative\\step time} \\
\midrule
\mzero{} & n/a & n/a & 2.0184 & 1.0000$\times$ & 162.6764 & 1.000$\times$ \\
$K=1$  & 1.754  & 1.850  & 3.604   & 1.7856$\times$  & 164.274 & 1.010$\times$ \\
$K=5$  & 7.840  & 7.526  & 15.366  & 7.6130$\times$  & 176.036 & 1.082$\times$ \\
$K=10$ & 16.732 & 16.688 & 33.420  & 16.5577$\times$ & 194.090 & 1.193$\times$ \\
$K=25$ & 35.221 & 35.637 & 70.858  & 35.1060$\times$ & 231.528 & 1.423$\times$ \\
$K=50$ & 76.385 & 74.277 & 150.662 & 74.6443$\times$ & 311.332 & 1.914$\times$ \\
\bottomrule
\end{tabular*}
\appendixtablenote{$M_0$ is the OPSD+SCGate baseline; each $K$ row uses \method{}. Teacher time is Pass 1 plus Pass 2. Both time ratios are normalized by OPSD+SCGate.}
\end{table}

Table~\ref{tab:step_timing_breakdown} gives the estimated step time breakdown at
$K=50$.

\begin{table}[!htbp]
\centering
\caption{Estimated training-step breakdown on Qwen3-8B at $K=50$.}
\label{tab:step_timing_breakdown}
\appendixtablestyle
\begin{tabular*}{\linewidth}{@{\extracolsep{\fill}}lrr@{}}
\toprule
Component & Time (s) & Share (\%) \\
\midrule
Rollout & 78.215 & 25.1227 \\
Teacher time (Pass 1 + Pass 2) & 150.662 & 48.3927 \\
Routing & 0.012 & 0.0039 \\
Student computation & 82.443 & 26.4807 \\
\midrule
Total & 311.332 & 100.0000 \\
\bottomrule
\end{tabular*}
\end{table}

\section{Additional Expert Analyses}
\label{app:expert_analyses}

\subsection{Full Reference-Side Audit}
\label{app:reference}

We audit the selected expert pools on the 500-question analysis set described
in Appendix~\ref{app:protocol}. Its complete reference solutions contain
331,246 token positions before filtering.

\begin{table}[!htbp]
\centering
\caption{Reference-side audit of the unperturbed teacher and selected expert pools.}
\label{tab:reference_full}
\appendixtablestyle
\begin{tabular*}{\linewidth}{@{\extracolsep{\fill}}crrrrrr@{}}
\toprule
\multirow{2}{*}{$\sigma$} & \multicolumn{2}{c}{SF coverage (\%)}
& \multicolumn{2}{c}{Token accuracy (\%)} & \multirow{2}{*}{JS($M_j,M_0$)}
& \multirow{2}{*}{\shortstack{MaxPeak\\peak}} \\
\cmidrule(lr){2-3}\cmidrule(lr){4-5}
& $M_0$ & Pool & $M_0$ top-1 & MaxPeak anchor & & \\
\midrule
.001 & 15.69 & 21.86 & 91.33 & 94.20 & .00111 & .9762 \\
.002 & 15.69 & 26.53 & 91.33 & 95.58 & .00393 & .9871 \\
.004 & 15.69 & 29.02 & 91.33 & 96.53 & .01198 & .9962 \\
.006 & 15.69 & 31.39 & 91.33 & 97.31 & .03474 & .9993 \\
\bottomrule
\end{tabular*}
\appendixtablenote{Selection-feasible (SF) coverage requires reference-side SCGate retention, positive PTA, and the selection clip. Accuracy, JS divergence, and peak probability use all SCGate-retained positions. $M_0$ is the unperturbed privileged teacher; Pool uses the selected perturbation experts.}
\end{table}

At every tested radius, the selected pool has higher selection-feasible
coverage than \mzero{}, and MaxPeak has higher anchor accuracy.
Larger radii also increase JS divergence and MaxPeak peak probability.

\subsection{Complete Raw MaxPeak Radius Probe}
\label{app:prefix}

\paragraph{Prefix construction.}
We generate trajectories on the \method{} training data using the frozen
Qwen3-8B base student. All trajectories use the same generation protocol.
After filtering and deduplication, we randomly sample 150 trajectories from
each of the following groups. The 300 trajectories come from 300 distinct
questions.
\begin{itemize}
    \item \texttt{complete\_wrong}: The trajectory contains
    \texttt{<|im\_end|>}, and its last boxed answer differs from the reference
    answer. We retain the first $75\%$ of its tokens as the continuation prefix.
    \item \texttt{truncated\_unknown}: The trajectory has no
    \texttt{<|im\_end|>}, usually because it reaches the 1,024-token generation
    limit. Its final correctness is unknown. We retain its complete token
    sequence as the continuation prefix.
\end{itemize}

\paragraph{Continuation evaluation.}
We evaluate each radius on the same 300 prefixes. Raw MaxPeak routes the
highest-peak expert at each decoding step and does not use the full router's
quantile selection. Both \mzero{} and Raw MaxPeak receive the same privileged
reference solution and may generate up to 4096 new tokens. This budget includes
reasoning tokens and excludes the student prefix and other input prompts.
A continuation is correct when its first closed boxed answer matches the
reference answer. Unfinished generations remain in the denominator. Completion
means producing a closed boxed answer within this budget. The FKL proxy uses
the union of retained top probabilities and one residual-mass bucket.

\begin{table}[!htbp]
\centering
\caption{Raw MaxPeak radius probe on 300 student-prefix continuations.}
\label{tab:prefix_full}
\appendixtablestyle
\begin{tabular*}{\linewidth}{@{\extracolsep{\fill}}crrrrrrr@{}}
\toprule
\multirow{2}{*}{$\sigma$} & \multicolumn{3}{c}{FKL proxy}
& \multicolumn{2}{c}{Accuracy (\%)} & \multicolumn{2}{c}{Completion (\%)} \\
\cmidrule(lr){2-4}\cmidrule(lr){5-6}\cmidrule(l){7-8}
& $M_0$ & Raw MP & Ratio & $M_0$ & Raw MP & $M_0$ & Raw MP \\
\midrule
.001 & .18138 & .18528 & 1.02$\times$ & 83.33 & 81.33 & 90.00 & 88.67 \\
.002 & .18138 & .20918 & 1.15$\times$ & 83.33 & \textbf{86.67}
& 90.00 & \textbf{94.67} \\
.004 & .18138 & .29848 & 1.65$\times$ & 83.33 & 62.33 & 90.00 & 79.00 \\
.006 & .18138 & .66341 & 3.66$\times$ & 83.33 & 27.33 & 90.00 & 39.33 \\
\bottomrule
\end{tabular*}
\appendixtablenote{Raw MP denotes Raw MaxPeak; the proxy ratio is Raw MP/$M_0$. All continuations are included.}
\end{table}

Raw MaxPeak gives the best continuation results at $\sigma=.002$. Compared
with \mzero{}, accuracy is $3.33$ points higher and completion is $4.67$ points
higher. At larger radii, the FKL proxy rises and continuation quality falls,
even though reference-side anchor accuracy continues to improve.

\subsection{Frequent Newly Covered Reference Tokens}
\label{app:tokens}

Table~\ref{tab:token_examples} lists frequent reference tokens at positions
newly covered by the perturbation pool at $\sigma=.002$. They include function
words such as \texttt{the} and \texttt{we}, LaTeX delimiters, numerals,
punctuation, and reasoning verbs such as \texttt{find} and \texttt{need}.
These are standard lexical and symbolic components of mathematical solutions.
The new coverage is therefore not limited to self-reflective style markers
such as \texttt{hmm} and \texttt{wait} \citep{zhao2026selfdistilled}.

\begin{table}[!htbp]
\centering
\caption{Frequent newly covered reference tokens at $\sigma=.002$.}
\label{tab:token_examples}
\appendixtablestyle
\begin{tabular*}{\linewidth}{@{\extracolsep{\fill}}lrr@{\hspace{18pt}}lrr@{}}
\toprule
Token & Approx. count & Share (\%) & Token & Approx. count & Share (\%) \\
\cmidrule(r){1-3}\cmidrule(l){4-6}
\texttt{the} & 406 & 4.70 & \texttt{)} & 107 & 1.24 \\
\texttt{\textbackslash} & 346 & 4.01 & \texttt{2} & 92 & 1.06 \\
\texttt{we} & 339 & 3.93 & \texttt{and} & 76 & .88 \\
\texttt{,} & 158 & 1.83 & \texttt{1} & 69 & .80 \\
\texttt{\textbackslash(} & 134 & 1.55 & \texttt{find} & 54 & .62 \\
\texttt{The} & 131 & 1.52 & \texttt{need} & 49 & .57 \\
\texttt{We} & 116 & 1.34 & & & \\
\bottomrule
\end{tabular*}
\appendixtablenote{Shares use an 8,638-token subset; approximate counts are derived from the shares and rounded to the nearest integer. Displayed tokens omit whitespace.}
\end{table}

\FloatBarrier
\section{Limitations and Future Work}
\label{app:limitations}

Our experiments cover mathematical reasoning tasks and the Qwen3 model family.
Evaluating \method{} on other model families and tasks, including code
generation, multi-hop question answering, and tool use, would test its broader
applicability.

Offline expert selection computes filtered PTA relative to the fixed
problem-only distribution $p_t^S$ on reference prefixes. Online routing
evaluates experts at states visited by the current student, but the selected
expert pool remains fixed throughout training. As the student changes, the
experts most useful to it may differ from those selected offline. Periodically
recomputing PTA and reselecting experts could reduce this mismatch. However,
repeated evaluation of all candidates would increase computational cost.
Future work can study efficient pool updates that balance this cost with gains
in student performance.

\end{document}